%% file: main.tex
\documentclass[acmsmall]{acmart}

\usepackage{todonotes}
\usepackage{graphicx}
\usepackage{hhline}
\usepackage{listings}
\usepackage{courier, times}
\usepackage{caption}
\usepackage{subcaption}
\DeclareGraphicsExtensions{.pdf,.jpg,.png}
\graphicspath{{figures/}}
\newcommand*{\thead}[1]{%
\multicolumn{1}{c}{\bfseries\begin{tabular}{@{}c@{}}#1\end{tabular}}}
\usepackage[framemethod=TikZ]{mdframed}

\mdfsetup{
    roundcorner=3pt,
    innerleftmargin=10pt,
    innerrightmargin=10pt,
    innertopmargin=5pt,
    innerbottommargin=10pt,
    frametitleaboveskip=5pt,
    skipabove=5pt,
    skipbelow=5pt
}

\mdfdefinestyle{greybox}{
    skipabove=10pt,
    skipbelow=20pt,
    middlelinecolor=gray!50,
    middlelinewidth=1.5pt,
    backgroundcolor=gray!5,
    frametitlefont={\normalfont\sffamily\bfseries}
}

\AtBeginDocument{%
  \providecommand\BibTeX{{%
    \normalfont B\kern-0.5em{\scshape i\kern-0.25em b}\kern-0.8em\TeX}}}

\setcopyright{acmcopyright}
\copyrightyear{2022}
\acmYear{2022}
\acmDOI{}

\begin{document}

\title[Tiny, Always-on and Fragile: Bias Propagation in On-device ML Workflows]{Tiny, Always-on and Fragile: Bias Propagation through Design Choices in On-device Machine Learning Workflows}

\author{Wiebke (Toussaint) Hutiri}
\authornote{The work was partially done while the author was an intern at Nokia Bell Labs.}
\email{w.toussaint@tudelft.nl}
\author{Aaron Yi Ding}
\affiliation{%
  \institution{Delft University of Technology}
  \streetaddress{Jaffalaan 5, 2628 BX}
  \city{Delft}
  \country{The Netherlands}
}

\author{Fahim Kawsar}
\author{Akhil Mathur}
\affiliation{%
  \institution{Nokia Bell Labs}
  \streetaddress{P.O. Box 1212}
  \city{Cambridge}
  \country{UK}
}
\renewcommand{\shortauthors}{Hutiri, et al.}

\begin{abstract}
Billions of distributed, heterogeneous and resource constrained IoT devices deploy on-device machine learning (ML) for private, fast and offline inference on personal data. On-device ML is highly context dependent, and sensitive to user, usage, hardware and environment attributes. This sensitivity and the propensity towards bias in ML makes it important to study bias in on-device settings. Our study is one of the first investigations of bias in this emerging domain, and lays important foundations for building fairer on-device ML. We apply a software engineering lens, investigating the propagation of bias through design choices in on-device ML workflows. We first identify \emph{reliability bias} as a source of unfairness and propose a measure to quantify it. We then conduct empirical experiments for a keyword spotting task to show how complex and interacting technical design choices amplify and propagate \emph{reliability bias}. Our results validate that design choices made during model training, like the sample rate and input feature type, and choices made to optimize models, like light-weight architectures, the pruning learning rate and pruning sparsity, can result in disparate predictive performance across male and female groups. Based on our findings we suggest low effort strategies for engineers to mitigate bias in on-device ML.

\end{abstract}



\newcommand{\parjump}{\vspace{+0.5em}}

\maketitle

\input{sections/1_introduction}
\input{sections/2_background}
\input{sections/3_overview}

\input{sections/4_bias_ondeviceml}
\input{sections/5_case_study}
\input{sections/6_results}

\input{sections/7_discussion}
\input{sections/8_conclusion}

\begin{acks}
This research is partially supported by SPATIAL project that has received funding from the European Union's Horizon 2020 research and innovation programme under grant agreement No.101021808. We thank Roel Dobbe, Sem Nouws and Dewant Katare for their feedback and useful suggestions on the work.
\end{acks}

\bibliographystyle{ACM-Reference-Format}
\bibliography{references, manual_references}

\clearpage
\appendix
\input{sections/appendix}

\end{document}

%% file: sections/1_introduction.tex
\section{Introduction}

Rising concerns about digital privacy and personal data protection~\cite{Naeini2017Privacy} are motivating a shift in data processing and machine learning (ML) from cloud servers to end devices~\cite{Chen2019Deep}. On-device ML is an emerging computing paradigm that makes this shift possible~\cite{banbury2020benchmarking}. In contrast to ML on centralized cloud-servers, on-device ML processes data directly on the device that collected them. This has important gains for privacy: if the data never leaves the device, the potential for unsolicited use or abuse by third parties is greatly reduced. Additionally, by eliminating data transfer during inference, on-device ML enables instantaneous, continuous and offline data processing, making it possible to operate devices in an always-on mode.

From earphones to embedded cameras, billions of tiny devices across the globe deploy on-device ML for inference on personal data. However, a growing body of research shows that ML systems are prone to bias~\cite{Mehrabi2019Survey, Pessach2022review}, and can lead to unfair predictions that favour or are prejudiced against particular groups of people. Bias in ML is concerning, as it can result in decisions that inflict harm on people, oftentimes vulnerable groups or minority populations~\cite{Birhane2022unseen}. Uber's fatal self-driving car crash in 2018~\cite{uber2019verge} acts as a stark reminder of the consequences of unchecked bias. The crash, which killed a pedestrian, was attributed to software design decisions which resulted in a series of misclassifications of the crash victim by Uber's ML system~\cite{uber2019arstechnica}. In the case of on-device ML, bias affects system reliability, that is to say the ability of on-device ML to "deliver stable and predictable performance in expected [operating] conditions"~\cite{nist2017framework}. Unexpected system failures can result in products and services that inflict harm on users or the public. To our knowledge, no prior studies have considered whether on-device ML systems are equally reliable for all users, that is to say, whether they are biased. 

\emph{In this paper we study bias in on-device ML. We approach the topic from a software engineering lens, and investigate the emergence of bias in the on-device ML workflow.} On-device ML presents a unique development environment. While the cloud offers limitless computing resources, on-device ML needs to account for the inherent hardware constraints of end devices: limited memory, compute and energy resources~\cite{Dhar2021Ondevice}. Developers aim to retain predictive accuracy while overcoming these constraints with engineering interventions in the on-device ML workflow. Engineering interventions demand that developers make design choices during product development. 

In previous work we investigated the impact of pre-processing parameter design choices in a keyword spotting task~\cite{Toussaint2021Characterising}. The study found that pre-processing parameters have a statistically significant impact not only on accuracy, but also on bias. The effect is more pronounced for light-weight neural network architectures and at lower sample rates, making this a relevant insight for the development of on-device ML. Informed by that work, we hypothesize that interdependent engineering design choices and unpredictable operating contexts can result in unexpected performance disparities between user groups in on-device ML applications. This paper builds on our previous study in three ways: 1) We develop a decision map of on-device ML to consider design choices in the development workflow systematically. 2) We expand our keyword spotting experiments on pre-processing parameters to more datasets and languages. 3) We conduct experiments to study bias due to design choices made during model compression.

Our paper is the first study of bias in on-device ML workflows, and makes the following contributions:
\begin{enumerate}
    \item We present a decision map to help developers identify design choices in the on-device ML workflow (\S\ref{s:ondevice}).
    \item We identify and quantify metrics to evaluate \emph{reliability bias} (\S\ref{s:framingbias}).
    \item  We empirically show that design choices made during model training (\S\ref{ss:model_training_choices}) and model optimization (\S\ref{ss:model_optimization_choices}) can amplify and propagate disparate performance across user groups, and thus \emph{reliability bias}.
    \item  We suggest strategies for mitigating \emph{reliability bias} without compromising accuracy (\S\ref{s:mitigation}).
\end{enumerate}

The paper starts with an overview of background knowledge and related work in \S\ref{s:fairnessinml}. We then present an overview of on-device ML and design choices arising during on-device ML development in \S\ref{s:ondevice}. In \S\ref{s:framingbias} we define and quantify \emph{reliability bias}. In \S\ref{s:casestudy} we introduce an empirical study of an audio keyword spotting task. We present our experimental results in \S\ref{s:results}, and propose strategies for mitigating \emph{reliability bias} in \S\ref{s:mitigation}. Finally we discuss the implications of our work for the development of fairer on-device ML in \S\ref{s:discussion} and conclude in \S\ref{s:conclusion}.

%% file: sections/2_background.tex
\section{Background and Related Work}
\label{s:fairnessinml}
Bias in ML software engineering is a new area of research. In this section we thus present interdisciplinary background literature on fairness and bias in ML. We define the concepts that we use in this work, and discuss bias measures and their limitations. We then highlight perspectives on bias in ML development from software engineering and statistical learning, with a focus on the impact of design choices.

\subsection{Bias and Fairness in Machine Learning}

\subsubsection{Concepts and Definitions}
Fairness in decision-making systems is considered as the "absence of any prejudice or favouritism toward an individual or a group based on their inherent or acquired characteristics"~\cite{Mehrabi2019Survey}. Biases in a system can render it unfair, and result in different individuals or groups of individuals being treated differently. If individuals belong to a protected class (e.g. based on their nationality, gender, socio-economic status or age) and they experience differential treatment that disadvantages them based on their membership in that class, this is considered discrimination and can carry legal consequences~\cite{DAlessandro2017conscientious}. An example of discrimination is denying an individual a home loan based on their gender. Bias can (but does not necessarily) lead to discrimination. We consider systems to be fairer and more inclusive if they are less biased. Practically, building ML systems that have no bias is difficult and maybe even impossible. However, quantifying and reducing bias are attainable and important steps towards building more inclusive and fairer ML systems.

\subsubsection{Bias Measures}
To measure bias, researchers in ML have quantified fairness measures that operationalize fairness definitions. Fairness definitions are categorized as measuring individual or group fairness~\cite{Mehrabi2019Survey}. Individual fairness measures require that similar people are treated similarly, while group fairness measures require that different groups are treated similarly. \citet{verma2018fairness} broadly categorise fairness measures into statistical (parity) measures, similarity-based measures and causal reasoning. Most statistical measures rely on metrics that calculate various ratios from error rates and prediction outcomes. A bias evaluation then establishes if metrics are equal for members of protected and unprotected groups. Equalised odds~\cite{Hardt2016equality}, for example, is a fairness measure that establishes if protected and unprotected groups have the same false negative and false positive error rates. 

Fairness measures can be further categorized as bias preserving and bias transforming, based on the measure's treatment of historic biases~\cite{wachter2021bias}. Parity-based fairness measures require equal error rates between groups of people. They are considered bias preserving as they propagate historic bias, for example through data labelling decisions which can replicate a biased world-view. \citet{wachter2021bias} argue that to support the objective of substantive equality in European non-discrimination law, fairness measures should be bias transforming. However, if labels can be exactly known, no historic bias exists, known performance disparities are legally justified or where systems are designed to replicate social bias, for example for the purpose of debugging, then bias preserving fairness measures are sufficient. In many on-device ML applications, such as wake-word detection, keyword spotting, object detection and speaker verification, data labels are exactly known and undisputed. We consider parity-based measures as useful measures of bias in this context. In \S\ref{s:framingbias} we introduce a parity-based bias measure that we use to quantify performance disparities between user groups in on-device ML. 

\subsection{Bias in the Machine Learning Workflow}

\subsubsection{Evidence of Bias in Decision-Making Systems}
The algorithmic fairness literature has focused predominantly on studying bias in ML systems for classification tasks, with a particular view towards the proliferation of decision-making systems that increasingly dominate public life~\cite{Mehrabi2019Survey}. Many studies have revealed evidence of bias in ML applications, ranging from natural language processing~\cite{Bolukbasi2016Man} and gender classification~\cite{Buolamwini2018Gender} to face recognition~\cite{Raji2020savingface} and automatic speech recognition~\cite{koenecke2020racial, Tatman2017Youtube}. As on-device ML is used for similar tasks, and leverages algorithmic approaches and data processing techniques from ML, it is necessary to investigate bias in on-device ML.

\subsubsection{Bias as a Concern for ML System Quality}

Recent work in software engineering has highlighted the need to model quality aspects of ML systems in detail~\cite{Siebert2021construction}. Bias has been identified as a new concern affecting ML software quality that should be considered as a non-functional requirement during development~\cite{Horkoff2019nonfunctional}. In requirements engineering and quality modeling, bias considerations are allocated to data-related aspects~\cite{Villamizar2022catalogue, Siebert2021construction}. However, from the perspective of statistical learning problems, bias can come from the training data, the predictive model and the evaluation mechanism~\cite{Mitchell2021Algorithmic}. The engineering and design nature of on-device ML requires an expanded view on bias to what is currently offered by the software engineering and statistical learning perspectives individually. Firstly, the bias of a component cannot be considered in isolation but must be considered within the evolving and dynamic system in which it is incorporated. Secondly, bias is not only a data concern. In reality, bias can emerge at different stages in the ML workflow and create reinforcing feedback loops~\cite{Sculley2015Hidden}. This paper expands current perspectives on bias in software engineering by studying how design decisions in the ML workflow influence bias. 

\subsubsection{Bias Propagation through Design Choices and System Composition}

Design choices play an important role in mitigating or propagating bias in the ML workflow. Mehrabi et. al~\cite{Mehrabi2019Survey} consider bias mitigation measures at different stages of the ML workflow: during pre-processing, in-processing and post-processing. This perspective aligns well with studies that consider the effects of design choices on ML bias in software engineering. For example, Hort and Sarro~\cite{Hort2022privileged} show how choosing thresholds for categorical data, a pre-processing decision, can impact both the degree of bias, and which groups are favoured. The Fair-SMOTE algorithm, on the other hand, is a pre-processing intervention that removes biased data labels and balances the training data distribution based on sensitive attributes and class labels~\cite{Chakraborty2021bias}. In on-device ML settings, trained ML models also undergo multiple post-processing steps to overcome resource constraints for on-device deployment and distribution shifts due to context heterogeneity. Some of these post-processing steps, like domain adaptation~\cite{Singh2021Fairness} and model compression~\cite{Hooker2020Characterising}, have been found to be biased. 

Holstein et. al~\cite{holstein2019improving} have observed that developers can feel a sense of unease at the societal impacts that their technical choices have, while Toussaint et. al~\cite{Toussaint2020design} have shown that early collaboration between clinical stakeholders and AI developers is important to guide design decisions to support social objectives within the public health sector. Dobbe et. al~\cite{Dobbe2021Hard} examine the impact of design choices on safety in AI systems for socio-technical decision-making in high-stakes social domains. Rather than looking at specific low-level technical choices in the ML workflow, they consider situations in which technical choices that promote different values are difficult to compare. They argue that developers ought to adopt diagnostic practices to proactively anticipate these choices and resolve them through feedback with stakeholders. Neglecting to do this, the authors further argue, gives rise to socio-technical gaps, where the technical functions do not satisfy the social requirements of AI systems. Drawing on these perspectives, this paper considers pre- and post-processing design decisions that arise in the inherently constrained on-device ML context, and examines the extent to which a relatively comprehensive set of design choices can support the social requirement for inclusive on-device ML.

%% file: sections/3_overview.tex
\section{On-Device Machine Learning Systems}
\label{s:ondevice}

Having laid the foundations for bias in ML, and the importance of design choices in propagating or mitigating bias in ML development, we turn our focus to on-device ML. Heterogeneous devices, diverse users and unknown usage environments make the performance of on-device ML highly context dependent. During development, engineers are faced with a large number of decisions to choose interventions that overcome hardware constraints and meet operational demands. Collectively, constraints and context-dependency make on-device ML development a complex engineering undertaking that requires mastery of hardware, software engineering and data processing techniques, alongside an in-depth understanding of the application context.  In this section we provide an overview of the data processing workflow for on-device ML systems, and highlight the various constraints, intervention strategies and design choices that an engineer encounters while designing on-device ML systems in practice. 

\subsection{Data Processing Workflow for On-device ML}
\label{ss:ondevice_development_workflow}

\begin{figure*}[thb]
    \centering
    \includegraphics[width=\textwidth]{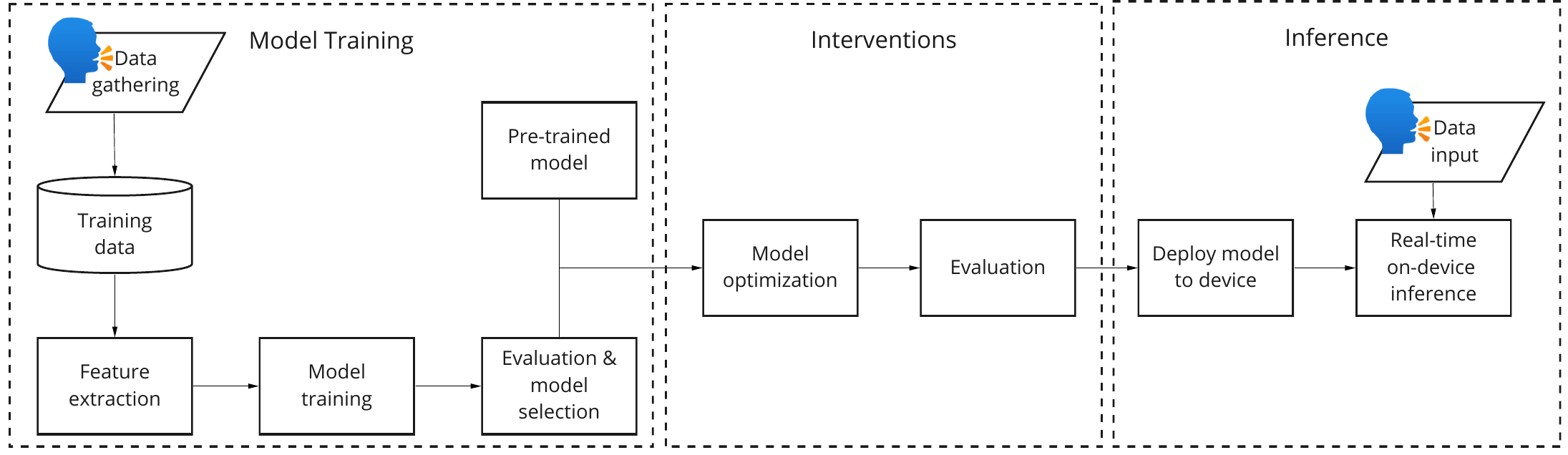}
    \caption{Data processing pipeline for on-device machine learning development}
    \label{fig:on_device_ml_pipeline}
\end{figure*}

The key processing steps during on-device ML development are model training, interventions, and inference. A typical data processing pipeline for on-device ML, as shown in Figure~\ref{fig:on_device_ml_pipeline}, consists of familiar ML processing steps for model training, evaluation, selection and inference. Key differences between on-device ML and cloud-based ML development arise due to the low compute, memory and power resources of end devices~\cite{Dhar2021Ondevice}. To enable on-device inference, interventions are needed to optimize a trained model and its data processing pipeline for on-device deployment. These aspects are described in greater detail below.

\subsubsection{Training}
The dominant approach for developing on-device ML is to delegate resource-intensive model training to the cloud, and to deploy trained and optimized models to devices~\cite{Dhar2021Ondevice}. The approach for training models is similar to typical ML pipelines: input data is gathered and undergoes a number of pre-processing operations to extract features from it. Thereafter, ML models are trained, evaluated and selected after optimizing a loss function on the data. Pre-trained models can also be downloaded and used if training data or training compute resources are not available. 

\subsubsection{Interventions}
The key differences between on-device ML and cloud-based ML development arise due to the low compute, memory and power resources of end devices~\cite{Dhar2021Ondevice}. To enable on-device deployment of the trained model, various \emph{interventions} are needed to optimize the model and its data processing pipeline. Common interventions include techniques such as model pruning, model quantization, or input scaling; all of which are aimed at optimizing device-specific performance metrics such as response time or latency~\cite{banbury2021mlperf}, memory consumption~\cite{Han2016deep}, or energy expenditure~\cite{Yang2018netadapt} with minimal impact on the model's accuracy. We elaborate on these intervention approaches in \S\ref{subsec.constraints}.

\subsubsection{Inference}
Once deployed, the trained and optimized model is used to make real-time, on-device predictions. On-device inference performance is determined by the model training process, from data collection to model selection, and the real-time sensor data input, but also by deployment constraints and interventions applied to the model. 

\subsection{Design Choices in On-device ML Engineering}
\label{subsec.constraints}
Building on the on-device data processing workflow that we described, we now discuss the key design choices that an engineer has to make in this workflow. We first explain the constraints of on-device ML that necessitate these design choices, and thereafter discuss the various interventions that can be taken to satisfy these constraints. We also highlight how these interventions could impact the accuracy and bias of on-device ML models. 

\subsubsection{Deployment Constraints}
On-device ML development needs to take into account the limited memory, compute and energy resources of the end devices~\cite{Dhar2021Ondevice}. The available \emph{storage and runtime memory} on a device limits the size of the ML models that can be deployed on it. The execution speed of inferences on the device is directly tied to the available \emph{compute} resources. Moreover, the amount of computations required by a model has a direct relation to its energy consumption; given that many end devices are battery powered with limited \emph{energy} resources, it becomes imperative that ML models operate within a reasonable energy budget. In addition to these resource constraints, on-device ML also has to deal with \emph{variations in the hardware and software stacks} of heterogeneous user devices~\cite{banbury2020benchmarking}. For instance, prior research~\cite{mathur2018using} has shown that different sensor-enabled devices can produce data at different sampling rates owing to their underlying sensor technology and real-time system state. Such variations can impact the quality of sensor data that is fed to the ML model, which in turn can impact its prediction performance.

\subsubsection{Interventions}
Research in on-device ML is largely concerned with overcoming these constraints and satisfying hardware-based performance metrics while achieving acceptable predictive performance~\cite{Dhar2021Ondevice}. Prior works have developed interventions to overcome \emph{memory} and \emph{compute} limitations, such as weight quantization~\cite{Han2016deep} and pruning~\cite{Liu2020pruning}. Other approaches such as input filtering and early exit~\cite{Huang2018multiscale}, partial execution and model partitioning~\cite{Dey2019EmbeddedPartitioning} allow for dynamic and conditional computation of the ML model depending on the available system resources. Another commonly used alternative to satisfy resource constraints is to design light-weight architectures that reduce the model footprint~\cite{Yang2018netadapt, Cai2020tinytl}. Finally, solutions have been proposed to make ML models robust to different resolutions of the input data~\cite{montanari2020eperceptive}, which is a key to dealing with sampling rate variations in end devices. Common to all these interventions is that they trade-off a model's resource efficiency with its prediction performance. For example, model pruning or the use of light-weight neural architectures can result in a model with smaller memory footprint and faster inference speed, however it comes at the expense of a slight accuracy degradation~\cite{Yang2018netadapt, Liu2020pruning, Cai2020tinytl}.    

\subsubsection{Design choices}
To build on-device ML, software engineers need to navigate deployment constraints and interventions alongside ML training and deployment. This is technically challenging, and charges engineers with the responsibility to take design actions and make design choices at each development step. Importantly, as on-device deployment constraints require interventions in the development process, design choices like the choice of model architecture, sample rate, input features and model compression techniques affect predictive and hardware performance, as well as bias~\cite{Toussaint2021Characterising}. Even though some design choices can be optimized through automated experimentation, iterating through all possible values requires extensive computing resources and time. This increases the cost of training. In countries and contexts where the low cost of on-device ML is a key enabler and driver for technology adoption~\cite{JanapaReddi2022WideningTinyML} increased training costs reduce technology access. Moreover, each design choice can introduce bias into the system. If time or compute are limited, engineers may need to limit the extent of their experimentation and only focus on a small set of choices.  

\begin{figure}[htb]
    \centering
    \includegraphics[width=0.95\textwidth]{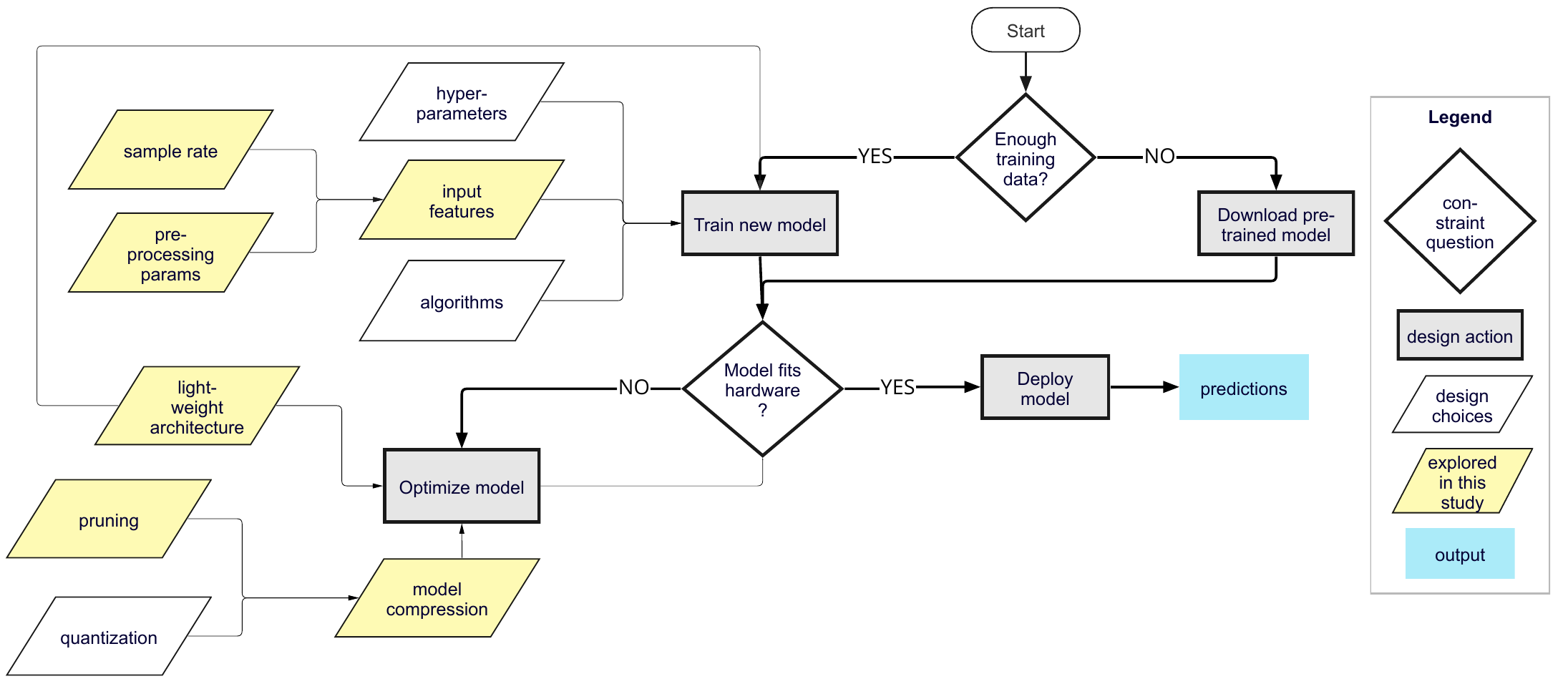}
    \caption{Decision map of design choices during on-device ML engineering. Yellow chart elements are design choices studied in this paper.}
    \label{fig:conceptual_model_design_choices}
\end{figure}

We visualize some of the key design choices as a decision map in Figure~\ref{fig:conceptual_model_design_choices}. The availability of training data is a logical starting point during development, as it determines whether a new model can be trained, or if a pre-trained model must be downloaded. Once an engineer commits to the design action of training a new model, they are confronted with design choices to select an algorithm, hyper-parameters, input features, pre-processing parameters and a data sample rate. After training or downloading the model, the engineer needs to determine if it fits within the memory, compute and power budget. If it does, they can deploy the model to make predictions. If the model does not fit within the hardware budget, the engineer must take design actions to optimize the model and reduce its resource requirements. This can be done through interventions like training a more light-weight architecture or compressing the model. These choices present further sub-choices, for example model compression can be done with pruning, quantization or both. In comparison to quantization, pruning involves more hyper-parameters and thus requires more design choices. Each design choice modifies the model, and has the potential of introducing bias in its predictions.

%% file: sections/4_bias_ondeviceml.tex
\section{Bias in On-device ML}
\label{s:framingbias}

On-device ML systems are deployed on billions of personal and low resource devices that continuously capture and monitor individuals or groups of people and the environment. In such cyber-physical systems (CPS) of distributed devices, ML functions mechanistically and is constructed from and activated by personal data collected with sensors. ML models can be seen as technical components of CPS, which along with other hardware and software components affect the system's functionality. Adopting a well-established definition\footnote{This definition has been adopted in the Cyper-Physical Systems Framework proposed by the National Institute of Standards and Technology in the United States} of reliability for trustworthy CPS~\cite{nist2017framework}, we posit that in order to be reliable and trustworthy, on-device ML should "deliver stable and predictable performance in expected [operating] conditions". Instead, if ML components fail unexpectedly, devices become unreliable and users can be inconvenienced or even harmed. 

Thus, we define an on-device ML model as biased if it causes devices to have disparate performance across user groups based on their personal or sensitive attributes. This can lead to devices that unexpectedly fail for some users, even if they deliver stable and predictable performance for others. If unexpected failures systematically affect particular users, these users are subjected disproportionately to harms that result from device failure. Biased on-device ML components thus lead to \emph{reliability bias} of a device, which then becomes a source of unfairness in the CPS. While a number of factors can lead to biased ML components, the focus of this work is to study how design choices during on-device ML development (see Figure~\ref{fig:conceptual_model_design_choices}) impact the model's accuracy for different demographic user groups, and thus propagate reliability bias.

\subsection{Quantifying Reliability Bias}
We consider an on-device ML model a reliable device component for a group if the group's predictive performance equals the model's overall predictive performance across all groups. If a model performs better or worse than average for a group, we consider it to be biased, showing favour for or prejudice against that group. Both favouritism and prejudice increase \emph{reliability bias}. We want to operationalize \emph{reliability bias} with a measure that captures these definitions and penalizes favouritism and prejudice equally. Additionally, the measure should be able to score models as being more or less biased, and should consider positive and negative prediction outcomes. Given these requirements, we first define bias of a model with respect to a group $i$ ($i = 1 \dots N$) as: 

\begin{equation}
\label{eq:sg_fairness}
    bias_{i} = ln \left(\frac{performance_{i}}{performance_{overall}}\right)
\end{equation}
where $performance_{i}$ is a metric computed for data samples belonging to the $i^{th}$ group, and $performance_{overall}$ is computed for all samples in the test set. $bias_{i}$ is 0 when a model is unbiased towards group $i$, negative when it performs worse than average and positive when it performs better than average for the group. The magnitude of the measure is equal for a performance ratio and it's inverse, as $ln(x) = -ln(\frac{1}{x})$. This has intuitive appeal that supports the interpretability of our measure: $bias_{i}$ is equal in magnitude but has opposing signs for groups that perform half as good and twice as good as average. Given the group bias scores, \emph{reliability bias} is the sum of absolute score values across all groups:
\begin{equation}
\label{eq:model_fairness}
    reliability\ bias = \sum_{i=1}^{N} \lvert bias_{i} \rvert
\end{equation}

In this paper we assume that all groups are equally important. The $reliability\ bias$ measure is thus unweighted and does not take group size into consideration. $Reliability\ bias$ has a lower bound of 0, and an infinite upper limit. Lower scores are preferred and signify that the performance across all groups is similar to the overall performance. We now turn towards an empirical audio keyword spotting (KWS) study to show how design choices in the on-device ML workflow propagate \emph{reliability bias}. 

%% file: sections/5_case_study.tex
\section{A Study on Bias in On-device Audio Keyword Spotting}
\label{s:casestudy}

In the remainder of the paper we examine the propagation of bias through design choices in on-device audio keyword spotting (KWS). KWS systems, which activate voice-based interactions with digital services~\cite{apple2021} on smart speakers and smart phones, are a prominent use case of on-device ML~\cite{banbury2021mlperf}. Voice-based service activation can be particularly beneficial for increasing access to digital services for individuals who suffer from restricted vision, mobility and movement, and for emergency response, home and elderly care. Many commercial products now exist that provide voice-activated urgent response with on-device KWS (e.g. "call help")~\cite{tuohy2021amazon, proknx2022}. Users place confidence in these products to support them in moments of crisis and provision them with access to critical care services. Despite the evident societal promise of on-device KWS, human speech signals exhibit variability based on social and physiological attributes of the speaker~\cite{hansen2015speaker}. This makes it essential to ensure that systems work reliably for all users irrespective of their personal attributes such as age and gender. A starting point for ensuring inclusive on-device audio KWS is to evaluate inference performance for speaker groups with different attributes. In this study we consider groups based on gender.

\begin{figure}[bht]
    \centering
    \includegraphics[width=0.7\textwidth]{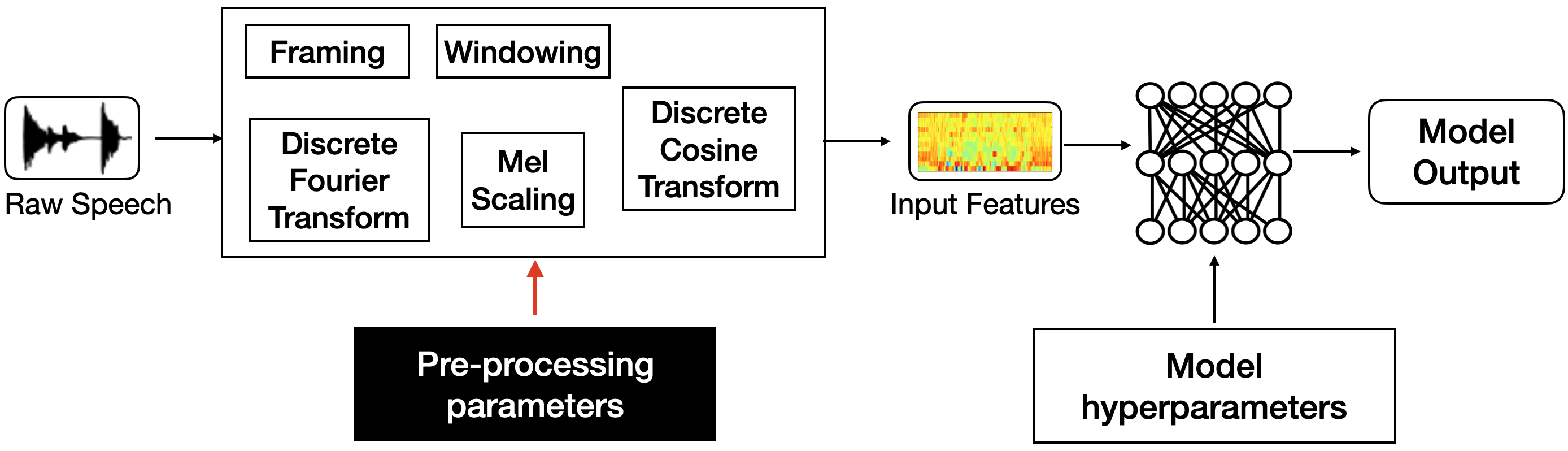}
    \caption{Audio processing pipeline during training and inference}
    \label{fig:audio_pipeline}
\end{figure}

\subsection{Overview of Audio Keyword Spotting Task}
\label{ss:audio_kws}

An audio keyword spotting system as shown in Figure~\ref{fig:audio_pipeline} takes a raw speech signal as input and outputs the keyword(s) present in the signal from a set of predefined keywords. Next we describe the end-to-end training and inference pipeline for a KWS system, while highlighting (\textbf{in bold}) the various design choices (ref. Figure~\ref{fig:conceptual_model_design_choices}) available to an ML engineer in this task. 

First, a raw speech signal is sampled from the microphone at a predefined \textbf{sample rate} (e.g., 8KHz, 16KHz) and split into overlapping, short time duration frames using a sliding window approach. This framing operation requires specifying a number of \textbf{pre-processing parameters}, which include i) \emph{Frame length} that defines the duration of each frame, ii) \emph{frame step} that indicates the step size by which the sliding window is moved, and iii) \emph{window function} which helps in reducing spectral leakage in Discrete Fourier Transform (DFT). Thereafter, each frame of the speech signal is transformed into \textbf{input features}: first, we apply a DFT to each frame to obtain log-scaled filter bank features known as \emph{log Mel spectrograms}. Optionally, log Mel spectrograms can be de-correlated using a Discrete Cosine Transform to generate Mel Frequency Cepstral Coefficiencts (\emph{MFCCs}). The number of log Mel spectrograms and MFCCs is also a designer-chosen parameter, often tuned empirically. Finally, the frame-level features (log Mel spectrograms or MFCCs) are concatenated across frames and mean-normalized to form a two-dimensional representation of the speech signal which is used to {train} a deep neural network classifier, as described in \cite{chen2014smallfootprint}. This process also involves choosing an appropriate \textbf{neural network architecture} that satisfies the resource constraints of the deployment device. Optionally, an ML engineer can also choose to optimize the trained neural network by applying various \textbf{model compression} techniques such as weight pruning (see \S\ref{subsec.constraints})

\subsection{Impact of Design Choices and Choice Variables}
\label{subsec.design-variables}
We now contextualize the design choices in KWS along the lines of the on-device ML workflow presented in \S\ref{s:ondevice}, and explain how prior literature has navigated them.

\parjump{}
\noindent
Firstly, the \emph{sample rate} can be seen as a deployment constraint due to hardware limitations such as microphone capabilities~\cite{montanari2020eperceptive}. Prior works~\cite{dieter2005power, small2021impact} have also used sample rate as a tunable parameter to adjust the power consumption on an embedded device during data collection. However, to our knowledge, there has been no study on how the choice of sample rate affects bias of ML models, other than our prior work~\cite{Toussaint2021Characterising} which is extended by this research.  

\parjump{}
\noindent
The choice of audio \emph{pre-processing parameters} is known to have an impact on the performance of a KWS model in an embedded system~\cite{montanari2020eperceptive}. Frame length and frame step together determine the temporal dimension of the 2D features that are fed to a DNN, and the number of log Mel spectrogram or MFCC features determine the length of features in each time segment. Together, these features influence the dimensions of the input data to the model, which in turn impacts the number of computations during inference. This insight was used in a recent work named ePerceptive~\cite{montanari2020eperceptive}, wherein the authors experimented with different values of \emph{frame step} to achieve a good trade-off between inference accuracy and latency. However, there has been no prior work which has explored potential accuracy-bias trade-offs due to {pre-processing parameters}. 

\parjump{}
\noindent
Unsurprisingly, the \emph{model architecture} plays an important role in the performance of a KWS system. 
Performing inferences on model architectures with fewer parameters takes less time, but could lead to accuracy degradation. On the contrary, deeper models with a large number of parameters might provide better accuracy, at the expense of higher inference latency. Prior KWS works~\cite{tucker2016model, he2017streaming, chen2014smallfootprint, alvarez2019endtoend, higuchi2020stacked, Zhang2017} have experimented with different architectures to achieve a good accuracy-latency trade-off. However none of these studies have  evaluated bias in KWS systems due to the choice of model architecture.

\parjump{}
\noindent
Finally, while compressing a trained model for deployment using model pruning, a developer needs to specify a number of parameters such as final sparsity, pruning frequency, schedule, and learning rate. \emph{Final sparsity}, specified in percentage, determines the proportion of weights that will be set to 0 during model pruning. Indeed, a high `final sparsity' leads to more compressed models, which result in lower storage requirements and reduced inference latency on the device~\cite{Liao2022empirical}. A \emph{pruning frequency} of `n' indicates that the model should be pruned after every `n' training steps. \emph{Pruning Schedule} can take two values: i) constant sparsity, which indicates the fixed sparsity level of the model throughout the training, or ii) polynomial decay, where the pruning sparsity grows rapidly in the beginning from initial\_sparsity, but then plateaus slowly to the final sparsity. Finally, \emph{Pruning Learning Rate} controls the step size taken by the model optimizer (e.g., Adam or Stochastic Gradient Descent) during backpropagation. Prior literature on neural network pruning primarily investigates the impact of final sparsity on model accuracy~\cite{Liao2022empirical} and does not shed light on the impact of other pruning parameters. However, given that these parameters also constitute important design decisions during model optimization, we choose to include them in our experiments. To our knowledge, the effect of pruning parameters on KWS models running on-device has not been studied.   

\subsection{Experiment Design}
\label{ss:experiment_design}

Having established that prior KWS literature has not adequately studied the impact of design choices on bias, we set up experiments to investigate design choices related to important design actions for on-device ML: model training and model optimization. Guided by the on-device ML development workflow presented in \S\ref{s:ondevice} and our prior work~\cite{Toussaint2021Characterising}, we aim to answer the following research questions within the context of an audio KWS task:

\begin{enumerate}
    \item How does the choice of architecture affect \emph{reliability bias} and accuracy?
    \item How does the audio sample rate affect \emph{reliability bias} and accuracy?
    \item How do pre-processing parameters affect \emph{reliability bias} and accuracy?
    \item How do pruning parameters affect \emph{reliability bias} and accuracy? 
\end{enumerate}

\parjump{}\noindent
The various design choices and choice variables are summarized in Table~\ref{tab:experiment_design} and explained next.

\begin{table}[hbt]
\small
    \centering
    \resizebox{\linewidth}{!}{
    \begin{tabular}{llll}
        \textbf{Design action} & \textbf{Design choice} & \textbf{Choice variable (unit)} & \textbf{Variable values} \\ \midrule
        Train new model & input features | sample rate & sample rate (kHz)  & 8, 16 \\ \noalign{\medskip}
        Train new model & input features | pre-processing & feature type & log Mel spectrogram, MFCC \\
        Train new model & input features | pre-processing & \# Mel filter banks & 20, 26, 32, 40, 60, 80 \\
        Train new model & input features | pre-processing & \# MFCCs & None, 10, 11, 12, 13, 14 \\
        Train new model & input features | pre-processing & frame length (ms) & 20, 25, 30, 40 \\
        Train new model & input features | pre-processing & frame step (\% frame length) & 40, 50, 60\\
        Train new model & input features | pre-processing & window function & Hamming, Hann\\ \noalign{\medskip}
        Optimize model & light-weight architecture & model architecture & CNN, low latency CNN~\cite{sainath2015convolutional} \\ \noalign{\medskip}
        Optimize model & model compression | pruning & final sparsity (\%) & 20, 50, 75, 80, 85, 90 \\
        Optimize model & model compression | pruning & pruning frequency & 10, 100\\
        Optimize model & model compression | pruning & pruning schedule & constant sparsity, polynomial decay\\
        Optimize model & model compression | pruning & pruning learning rate & 1e-3, 1e-4, 1e-5\\ \noalign{\medskip}
    \end{tabular}}
    \caption{Overview of design choice variables and values for the audio keyword spotting study}
    \label{tab:experiment_design}
\end{table}

\parjump{}\noindent
As discussed earlier, the \emph{neural network architecture} is an important design choice during model training. We experiment with two convolutional neural network (CNN) architectures for KWS, originally proposed in~\cite{sainath2015convolutional} and later implemented in the TensorFlow framework. The architecture that we refer to as \emph{CNN} consists of two convolutional layers followed by one dense hidden layer, while the low-latency CNN (\emph{llCNN}) consists of one convolution layer followed by two dense hidden layers. The authors in \cite{sainath2015convolutional} showed that the llCNN architecture, by virtue of having less convolution operations, is more optimized for on-device KWS. 

\parjump{}\noindent
Next, we study choices that affect the input features of the model, namely \emph{sample rate} and \emph{pre-processing parameters}. Audio keyword spotting developer benchmarks often use a 16kHz audio input~\cite{warden2018speech,mazumder2021multilingual}. In practice many devices collect data at a lower sample rate of 8kHz~\cite{montanari2020eperceptive} due to hardware constraints. We thus train models with audio data at two sample rates, 16kHz and 8kHz, for both architectures. 

\parjump{}\noindent
For studying the impact of pre-processing parameters, we take inspiration from prior KWS literature~\cite{tucker2016model, he2017streaming, chen2014smallfootprint, alvarez2019endtoend, higuchi2020stacked, Zhang2017}, and experiment with two feature types, log Mel spectrograms and MFCCs. More specifically, we vary the dimensionality of log Mel spectrograms from 20 to 80, and of MFCCs from 10 to 14. We also consider log Mel spectrograms that are used directly as input features, with no MFCCs. Further, we experiment with three temporal pre-processing parameters: frame length (20-40 ms), frame step (40\%-60\% overlap) and the window type (Hamming/Hann); these values are based on prior on-device KWS works~\cite{he2017streaming, alvarez2019endtoend, higuchi2020stacked, Zhang2017, montanari2020eperceptive}.  

\parjump{}\noindent
With regards to model optimization, we focus on model compression, in particular \emph{parameter choices during post-training pruning}. Based on prior literature~\cite{liebenwein2021lost, Liao2022empirical}, we vary the pruning sparsity from 20\% to 90\%. For pruning schedule, we experiment with both constant sparsity and polynomial decay as explained in \S\ref{subsec.design-variables}. For learning rate, we choose three values based on prior KWS literature on model training~\cite{tucker2016model, he2017streaming, chen2014smallfootprint, alvarez2019endtoend, higuchi2020stacked, Zhang2017}. For pruning frequency, we used two values: 100 (the default frequency in TensorFlow) and a faster option of 10, wherein the pruning operation takes place after every 10 training steps.  

\subsection{Experiment Setup}

\subsubsection{Datasets}

We trained and evaluated models on the following five spoken keywords datasets spanning four languages --- English, German, French, and Kinyarwanda: 

\parjump{}
\noindent
\textbf{Google Speech Commands} (\emph{google\_sc})~\cite{warden2018speech} is an English language dataset consisting of 104,541 spoken keywords from 35 keyword classes such as \emph{Yes, No, One, Two, Three}, recorded by volunteer contributors and released at a 16kHz sample rate. We labeled every utterance as male or female using a crowd-sourced data labelling campaign conducted on Amazon Mechanical Turk. We used the same train, validation and test set splits of 85\%, 10\%, 5\% respectively from the original dataset. Female speakers constituted 30\% of the original training data, 32\% of the validation and 29\% of the test data. During training we thus ensured that mini-batches have an equal balance of male and female speakers by randomly sampling from the male training set. 

\parjump{}
\noindent
\textbf{Multilingual Spoken Words Corpus Datasets}~\cite{mazumder2021multilingual}. The Multilingual Spoken Words Corpus (MSWC) is a large corpus of spoken
words in 50 languages, originally sampled at 48KHz. Each language partition contains hundreds of hours of audio data with tens of thousands of keyword classes. MSWC has been derived from Mozilla Common Voice\footnote{\url{https://commonvoice.mozilla.org/}} by splitting the crowd-sourced, read-speech corpus into individual words. We chose four of the languages with the largest data resources in MSWC to create four keyword spotting datasets in different languages: \textbf{MSWC English} (\emph{mswc\_en}), \textbf{MSWC German} (\emph{mswc\_de}), \textbf{MSWC French} (\emph{mswc\_fr}) and \textbf{MSWC Kinyarwanda} (\emph{mswc\_rw}). Each of the MSWC datasets was created with data from its language partition, and a consistent approach to select keywords, balance data across male and female speakers, and split the dataset into train, validation and test splits.

For each dataset, we selected keywords from the 35 largest keyword classes to create training datasets that are equivalent to Google Speech Commands. Following the keyword selection strategy of the authors of the MSWC dataset, we only selected keywords with more than 3 characters. Additionally, if two words started with the same 3 letters, we only selected the first occurring word. This resulted in a total of 200 628 keyword utterances for MSWC English, 85 572 keyword utterances for MSWC German, 75 644 keyword utterances for MSWC French and 53 608 keyword utterances for MSWC Kinyarwanda. The dataset sizes vary based on the language representation in the Mozilla Common Voice corpus.

\begin{table}[hbt]
    \centering
    \resizebox{\linewidth}{!}{
    \begin{tabular}{c|cccc}
        dataset split & \textbf{MSWC English} & \textbf{MSWC German} & \textbf{MSWC French} & \textbf{MSWC Kinyarwanda} \\ \midrule
        female training & 79002 (39\%)& 34728 (41\%) & 31127 (41\%) & 20713 (39\%)\\
        male training & 79611 (40\%) & 34329 (40\%) & 30276 (40\%) & 21786 (41\%) \\
        female validation & 10496 (5.2\%) & 4613 (5.4\%) & 2790 (3.7\%) & 3580 (6.7\%) \\
        male validation & 10238 (5.1\%) & 3976 (4.6\%) & 3601 (4.8\%) & 1801 (3.4\%) \\
        female test & 10816 (5.4\%) & 3445 (4\%) & 3905 (5.2\%) & 2511 (4.7\%) \\
        male test & 10465 (5.2\%) & 4481 (5.2\%) & 3945 (\%) & 3217 (6\%) \\ \midrule
        total & 200628 & 85572 & 75644 & 53608
    \end{tabular}}
    \caption{Audio keyword utterance count (and \% of total dataset) across dataset splits for MSWC English, MSWC German, MSWC French and MSWC Kinyarwanda datasets.}
    \label{tab:mswc_splits}
\end{table}

To ensure gender-balanced datasets, we only used keyword utterances where the gender metadata field was male or female. The gender metadata in Mozilla Common Voice has been provided by data donors and thus corresponds with the self-identified gender of the speaker. For each keyword we counted the utterances per gender. We included all utterances/keyword of the gender with fewer utterances/keyword, and randomly sampled the same number of utterances/keyword from the gender with more utterances/keyword. We joined the selected data for both genders and all keywords, before splitting the data into train, validation and test sets. To create the dataset splits, we followed the protocol described in~\cite{mazumder2021multilingual} as closely as possible while enforcing gender-balance. We first created a list of unique keyword-speaker pairs so that train, validation and test sets are separate. Next, we randomly sampled 80\% of keyword-speaker pairs for training. We then randomly sample 10\% of keyword-speaker pairs for validation, excluding pairs already in the training set and rounding to the nearest integer. Finally, we allocated the remaining keyword-speaker pairs to the test set. The keyword utterance count and representation for male and female genders across dataset splits are shown in Table~\ref{tab:mswc_splits}. During training we ensured that mini-batches have an equal balance of male and female speakers. 

\subsubsection{Training Details}
\label{ss:training_details}
Our training setup is implemented in Tensorflow 2.0 and we used a Nvidia V100  GPU to train the models. For each dataset we iteratively trained models with all combinations of model architectures, sample rates and pre-processing parameters listed in Table~\ref{tab:experiment_design}. This resulted in 3456 candidate models per dataset, and a total of 17280 experiments across 5 datasets. We used the TF HParams API\footnote{\url{https://www.tensorflow.org/tensorboard/hyperparameter_tuning_with_hparams}} for tuning the training-time learning rate for each model from the following three options: \{1e-2, 1e-3, 1e-4\}. We used the Adam optimizer for training, with a fixed batch size of 128 samples. Each model was trained for 10 epochs, which was chosen based on empirical evidence that the model performance did not improve beyond 10 epochs. 

Thereafter, we used model selection criteria that consider accuracy and bias (discussed in detail in \S\ref{ss:model_selection}) to select baseline models for model compression. Table~\ref{tab:pruning_experiments} lists the number of baseline models selected per dataset for the pruning experiments. For the Google SC and MSWC Kinyarwanda datasets we could not find models that met all our selection criteria across architectures and sample rates, which is why fewer baseline models were selected for these datasets. We then obtained the compressed version of the baseline models under each combination of pruning parameters listed in Table~\ref{tab:experiment_design}. As with training, we also used 10 epochs for pruning. Our pruning experiments resulted in 72 pruned models for each baseline model, and a total of 12168 experiments.

\setlength{\tabcolsep}{5pt}
\begin{table}[hbt]
\small
    \centering
    \begin{tabular}{l|ccccc}
         & \thead{Google\\SC} & \thead{MSWC\\German} & \thead{MSWC\\English} & \thead{MSWC\\French} & \thead{MSWC\\Kinyarwanda}\\ \midrule
        \textbf{16kHz CNN} & 9 & 9 & 9 & 9 & 6\\
        \textbf{16kHz llCNN} & 8 & 9 & 9 & 9 & 7 \\
        \textbf{8kHz CNN} & 9 & 9 & 9 & 9 & 7 \\
        \textbf{8kHz llCNN} & 9 & 9 & 9 & 9 & 6\\
    \end{tabular}
    \caption{Number of baseline models pruned per dataset, architecture and sample rate}
    \label{tab:pruning_experiments}
\end{table}

\subsubsection{Evaluation Protocol}

As ground truth labels in audio KWS can be exactly known and are unambiguous, we assume that labels are correct and unbiased. We can thus evaluate bias with the parity-based \emph{reliability bias} measure which we defined in Equation~\ref{eq:model_fairness}. For our experiments we compute two evaluation metrics for each model on the held-out test set: i) reliability bias and ii) accuracy. For accuracy, we compared five different metrics: Cohen's kappa coefficient, precision, recall, weighted F1 score and the Matthews Correlation Coefficient (MCC). The trends we observed are consistent across metrics. Thus, we only report accuracy results for the MCC, which is a robust metric for multiclass classification. 

%% file: sections/6_results.tex
\section{Empirical Results and Analysis}
\label{s:results}
In this section we present the results of our study and analyze the impact of design choices on reliability bias and accuracy during different stages of the on-device audio KWS workflow. We start with design choices that arise during model training, first analyzing the impact of the architecture and sample rate, then the impact of pre-processing parameters. After that we consider model optimization design choices that arise during model compression, namely the impact of pruning hyper-parameters. 

\subsection{Design Choices during Model Training}
\label{ss:model_training_choices}

To analyze the impact of the pre-processing parameters, we performed factorial ANOVA tests that allow for interactions~\cite{Navarro2019learning} on our balanced study design. This type of statistical test is used to determine the influence of two or more independent variables on one dependent variable, which makes it suitable for our study. We coded deviation (or sum) contrasts and used type~3 sums of squares. The analysis was done in python using the $scipy\ stastmodels$ package and is available as a jupyter notebook on github~\footnote{\url{https://github.com/akhilmathurs/fair-ondevice-ML}}. Given the large number of possible interactions between our independent variables (i.e. choice variables in Table~\ref{tab:experiment_design}), we designed the first factorial ANOVA model (see Model~\ref{mod:anova1} in the Appendix) to consider a subset of interactions that we deemed important for accuracy and reliability bias of KWS models based on prior visual analysis. We continued to improve the factorial ANOVA models separately for the two dependent variables (MCC (accuracy) and reliability bias) by removing all non-significant interactions, and then including lower-level interactions. The final ANOVA models are included in the Appendix, with Models~\ref{mod:anova2} and~\ref{mod:anova3} capturing variables and interactions of model training design choices on the accuracy score and reliability bias respectively.

Tables~\ref{tab:parameter_importance_mcc} and~\ref{tab:parameter_importance_bias} show statistically significant interaction and main effects of the final factorial ANOVA models. For completeness we have included main effects even if they already contribute to an interaction. The final factorial ANOVA models are significant (MCC~(accuracy): $F(171)=2527.2$, $p=0.0$, $R^2_{adj.}=0.9615$ and reliability bias: $F(92)=367.95$, $p=0.0$, $R^2_{adj.}=0.6615$). For reference, the critical F statistics at p-values less than 0.01 and 0.05 are shown in Table~\ref{tab:fcrit_at_p}. Based on the F statistics, we reject the null hypothesis that neither design choices made during model training, nor their interactions affect KWS model accuracy and reliability bias. The $R^2$ values indicate that the accuracy ANOVA model ($R^2=0.9619$, $R^2_{adjusted}=0.9615$) better captures the effects than the reliability bias ANOVA model ($R^2=0.6633$, $R^2_{adjusted}=0.6615$), in which a portion of variance in the dependent variable remains unaccounted for. Next we examine the impact of the model architecture and sample rate, and of the pre-processing parameters in greater detail.

\setlength{\tabcolsep}{10pt}
\renewcommand{\arraystretch}{1.1}
\begin{table}[hbt]
\centering
\resizebox{\linewidth}{!}{
\begin{tabular}{lcccc}
\toprule
\textbf{Factorial ANOVA main and interaction effects} & \textbf{SS} & \textbf{df} & \textbf{F} & \textbf{p(\textless{}0.05)} \\ \midrule
model architecture & 31.9714 & 1 & 6.0103E+04 & 0.0E+00 \\
sample rate & 3.4011 & 1 & 6.3938E+03 & 0.0E+00 \\
dataset & 160.1572 & 4 & 7.5270E+04 & 0.0E+00 \\
mfccs & 17.2272 & 5 & 6.4771E+03 & 0.0E+00 \\
mel filter banks & 3.1283 & 5 & 1.1762E+03 & 0.0E+00 \\
frame step & 0.1500 & 2 & 1.4103E+02 & 1.8E-61 \\
model architecture * mel filter banks & 0.0202 & 5 & 7.6075E+00 & 3.8E-07 \\
dataset * sample rate * mfccs & 0.1425 & 20 & 1.3391E+01 & 7.0E-45 \\
dataset * model architecture * mfccs & 0.1056 & 20 & 9.9288E+00 & 3.5E-31 \\ 
Residual & 9.100465 & 17108 & - & - \\\midrule
\multicolumn{1}{r}{\textbf{Model}} & - & 171 & 2.5277E+03 & 0.0E+00 \\
\multicolumn{1}{r}{\textbf{$R^{2}$:}} & 0.9619 &  &  &  \\
\multicolumn{1}{r}{\textbf{Adjusted $R^{2}$:}} & 0.9615 &  &  & \\ \hhline{=====}
\end{tabular}}
\smallskip
\caption{Significant main and interaction effects of model training design choices on \textbf{MCC (accuracy)}. SS=sum of squares, df=degrees of freedom}
\label{tab:parameter_importance_mcc}
\vspace{-2em}
\end{table}

\setlength{\tabcolsep}{10pt}
\renewcommand{\arraystretch}{1.1}
\begin{table}[hbt]
\centering
\resizebox{\linewidth}{!}{
\begin{tabular}{lcccc}
\toprule
\textbf{Factorial ANOVA main and interaction effects} & \textbf{SS} & \textbf{df} & \textbf{F} & \textbf{p(\textless{}0.05)} \\
\midrule
model architecture & 0.96734 & 1 & 469.248554 & 1.10E-102 \\
sample rate & 0.477009 & 1 & 231.393075 & 6.45E-52 \\
dataset & 62.4386 & 4 & 7.5721E+03 & 0.0E+00 \\
mel filter banks & 0.1225 & 5 & 1.1887E+01 & 1.7E-11 \\
dataset * sample rate & 0.7840 & 4 & 9.5078E+01 & 3.9E-80 \\
dataset * mel filter banks & 0.3758 & 20 & 9.1140E+00 & 5.1E-28 \\
dataset * model architecture * mfccs & 0.9662 & 20 & 2.3436E+01 & 1.8E-85 \\
Residual & 35.4366 & 17190 & - & - \\
\midrule
\multicolumn{1}{r}{\textbf{Model}} & - & 92 & 3.6795E+02 & 0.0E+00 \\
\multicolumn{1}{r}{\textbf{$R^{2}$:}} & 0.6633 &  &  & \\
\multicolumn{1}{r}{\textbf{Adjusted $R^{2}$:}} & 0.6615 &  &  & \\ \hhline{=====}
\end{tabular}}
\smallskip
\caption{Significant main and interaction effects of model training design choices on \textbf{reliability bias}. SS=sum of squares, df=degrees of freedom}
\label{tab:parameter_importance_bias}
\vspace{-2em}
\end{table}

\begin{table}[hbt]
\resizebox{\linewidth}{!}{
\centering
\begin{tabular}{c|cccccccccc}
df & 1 & 2 & 4 & 5 & 8 & 10 & 20 & 40 \\\midrule
$F_{crit}$ (p\textless{}0.01) & 4052.1807 & 98.5025 & 21.1977 & 16.2582 & 11.2586 & 10.0443 & 8.0960 & 7.3141 \\
$F_{crit}$ (p\textless{}0.05) & 161.4476 & 18.5128 & 7.7086 & 6.6079 & 5.3177 & 4.9646 & 4.3512 & 4.0847
\end{tabular}}
\smallskip
\caption{Critical F-values for determining significance at p\textless{}0.01 and p\textless{}0.05 for different degrees of freedom (df)}
\label{tab:fcrit_at_p}
\end{table}

\subsubsection{Impact of Model Architecture and Sample Rate}
\label{ss:impact_of_sample_rate}

The results of the statistical tests in Tables~\ref{tab:parameter_importance_mcc}~and~\ref{tab:parameter_importance_bias} show that model architecture and sample rate contribute to significant interaction effects that impact accuracy and reliability bias. We now examine how these two metrics are affected by variable values and their interactions. 

\begin{figure}[hbt]
    \centering
    \includegraphics[width=\textwidth]{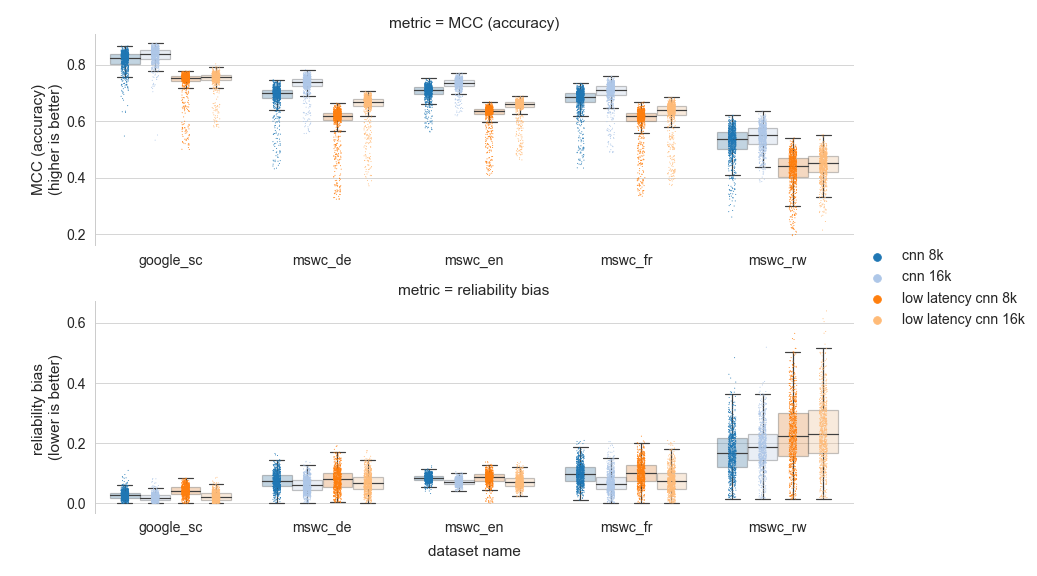}
    \caption{Experimental results of MCC (accuracy) and \emph{reliability bias} for CNN and low latency CNN model architectures with 16kHz and 8kHz sample rates trained on 5 different datasets.}
\label{fig:accuracy_bias_experiments}
\end{figure}

Figure~\ref{fig:accuracy_bias_experiments} shows a boxplot of accuracy and reliability bias for CNN and light-weight low latency CNN (llCNN) architectures trained on 16kHz and 8kHz audio data. A higher MCC score implies better prediction performance. The trends in accuracy scores for models trained with different architectures and sample rates are consistent across datasets. CNN and llCNN architectures trained at 8kHz have a lower median accuracy score (i.e. they are worse) than those trained at 16kHz, and CNN architectures have higher scores than their light-weight counterparts. While models trained on the mswc\_rw dataset still follow this trend, their performance, in general, is considerably worse than that of the other models. Possible reasons for this are that less training data was available for these models, and Kinyarwanda is a different language family than the languages in the other datasets. It is out of the scope of this study to consider bias due to language and accent, which remains an important area for future work.

For reliability bias we observe that median scores are higher (i.e. worse) for models trained at 8kHz than those for models trained at 16kHz. For the google\_sc, the mswc\_de and the mswc\_fr datasets, models trained at lower sample rates also have a higher interquartile range (IQR) in reliability bias scores. The light-weight llCNN architecture tends to have a higher median reliability bias and greater IQR than the CNN architecture, but the effect is not as pronounced as for accuracy. Models trained on the mswc\_rw dataset do not follow these trends. While median reliability bias of CNN models is lower than that of llCNN models, 8kHz models are also less biased than 16kHz models. We anticipate that the deviation between trends observed for the mswc\_rw models and the remaining models contributes significantly to the large effect size of the dataset variable that we observe in Tables~\ref{tab:parameter_importance_mcc}~and~\ref{tab:parameter_importance_bias}.

\begin{figure}[tb]
        \centering
        \includegraphics[width=\textwidth]{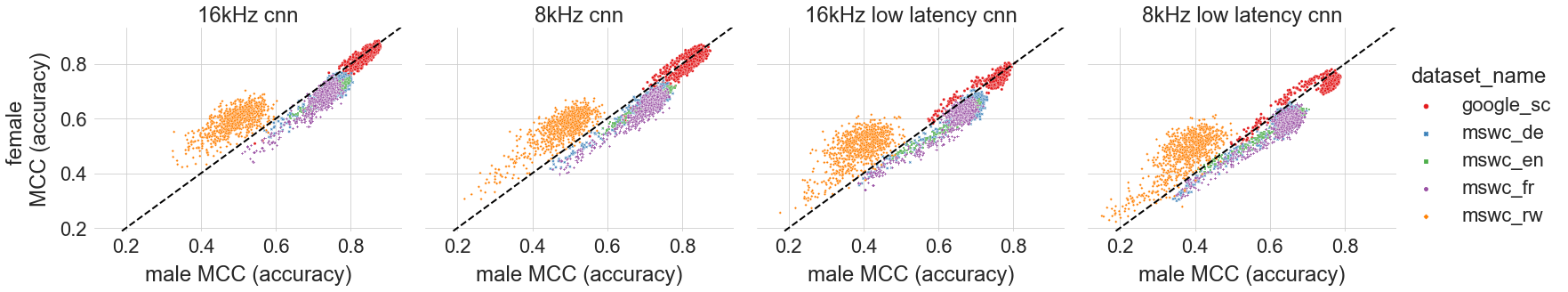}
    \caption{Disaggregated MCC (accuracy) scores for males (x-axis) and females (y-axis) for a single model trained with a unique combination of pre-processing parameters. On the black diagonal the performance is equal for both subgroups.}
\label{fig:subgroup_performance_accuracy}
\end{figure}

Delving deeper into these findings, we analyze the relationship between male and female MCC scores across architectures and sample rates in Figure~\ref{fig:subgroup_performance_accuracy}. Each data point represents the disaggregated male and female accuracy scores of a single model trained with a unique combination of pre-processing parameters. The dotted black diagonal represents equal performance for male and female speakers. Points above the diagonal perform better for females, and points below perform better for males. For the mswc\_de, \_en and \_fr datasets it is evident that accuracy scores are biased to favour male speakers. For the mswc\_rw dataset, models always favour female speakers. For the google\_sc dataset the results are more nuanced. Models trained with CNN architectures tend to favour male speakers, whereas models trained with llCNN tend to favor female speakers. This figure shows that the nature of the training data contributes significantly to bias. We also observe that for each dataset there exist models that lie on or very close to the diagonal. These models have a lower reliability bias than the remaining models. We hypothesize that pre-processing parameters contribute to reliability bias, and thus the distance of experiments from the diagonal. This leads us to the next section, where we analyze the role of pre-processing parameters on accuracy and reliability bias. 

\begin{mdframed}[style=greybox] \textbf{Key insights}: Model accuracy is lower at lower sample rates and for light-weight architectures. Median and IQR of reliability bias tend to be greater at lower sample rates and for light-weight architectures. The direction of bias is strongly influenced by the training dataset. Overall, male speakers are favoured by models. An exception to this are models trained on the mswc\_rw dataset, which have considerably lower accuracy and favour female speakers.
\end{mdframed}

\subsubsection{Impact of Pre-processing Parameters}
\label{ss:impact_of_preprocessing_parameters}
\begin{figure}[t]
        \centering
        \includegraphics[width=\textwidth]{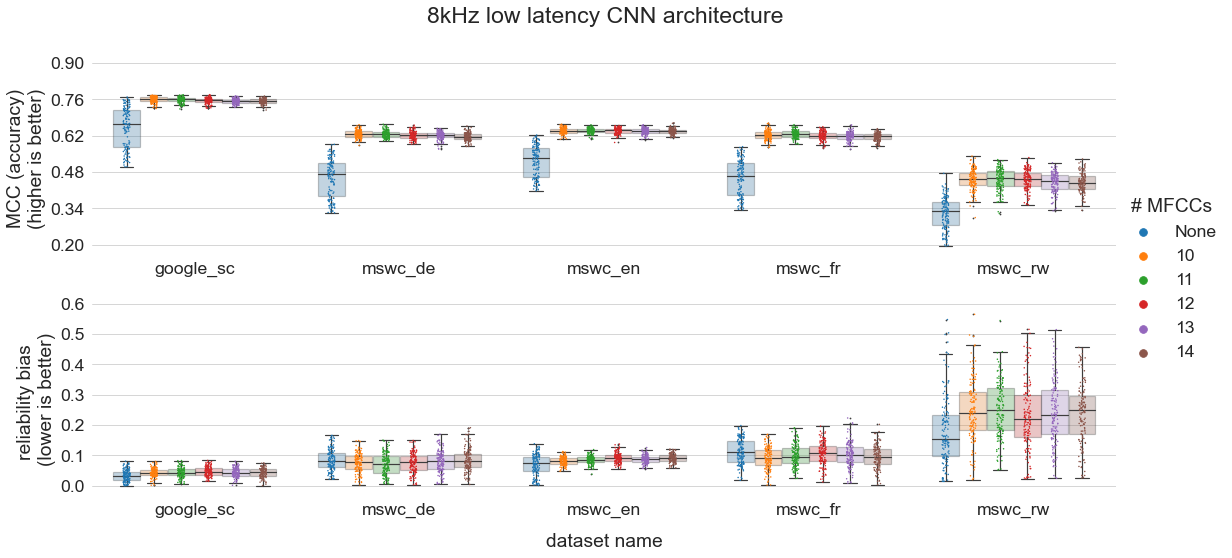}
    \caption{Effect of MFCC dimensions on accuracy and reliability bias for \textbf{8kHz low latency CNN} models. Models without MFCC features (blue), i.e. models that directly use log Mel spectrograms as input features, perform considerably worse than those that use MFCC features.}
\label{fig:dataset_arch_mfccs_8}
\end{figure}

Having studied the effect of the architecture and sample rate, we now turn to pre-processing parameters, the next design choice listed in Table~\ref{tab:experiment_design}. The F statistics and p-values in Tables~\ref{tab:parameter_importance_mcc} and~\ref{tab:parameter_importance_bias} indicate that the dimensions of log Mel spectrograms and MFCC features significantly affect accuracy and reliability bias. For accuracy there exist interaction effects between Mel filter banks and architecture, between MFCCs, dataset and sample rate, and between MFCCs, dataset and architecture. The latter interaction effect also exists for reliability bias, as well as an interaction effect between Mel filter banks and dataset. Figure~\ref{fig:dataset_arch_mfccs_8} visualizes accuracy and reliability bias for the six MFCC and log Mel spectrogram dimensions across all datasets for the 8kHz low latency CNN models. As highlighted earlier, the lower sample rate and light-weight architecture result in models that experience greater decline in accuracy and reliability bias. We thus anticipate that the impact of pre-processing parameters is more pronounced for these models. 

In Figure~\ref{fig:dataset_arch_mfccs_8} the number of MFCC dimensions implies the choice of input feature type. Models with no MFCCs (i.e. \# MFCCs = None) use only log Mel spectrograms as input features. It is clear from the figure that the accuracy of models trained with log Mel spectrograms (i.e. the blue boxes) is significantly worse than that of models trained with MFCC input features. For models trained with MFCC features, fewer dimensions (i.e. \# MFCCs = 10 or 11) tend to result in a higher median accuracy than more dimensions. However, the impact of this is much smaller than that of using log Mel spectrograms. For reliability bias we observe mixed results that depend on the training dataset. Google\_sc, mswc\_en and mswc\_rw have a lower median reliability bias when using log Mel spectrograms. On the other hand, for the mswc\_de and mswc\_fr datasets the median reliability bias is lower for models trained with MFCC input features. Figure~\ref{fig:dataset_arch_mfccs_16} in the Appendix shows comparable results for 16kHz CNN models. Here we still observe that the median accuracy is lower for log Mel spectrogram input features, except for the google\_sc dataset. This dataset also has a lower median and smaller IQR of reliability bias scores when using log Mel spectrograms. Overall, the impact of the number of MFCC dimensions and by association the input feature type is less pronounced for CNN models trained at 16kHz.    

\begin{figure}[t]
        \centering
        \includegraphics[width=\textwidth]{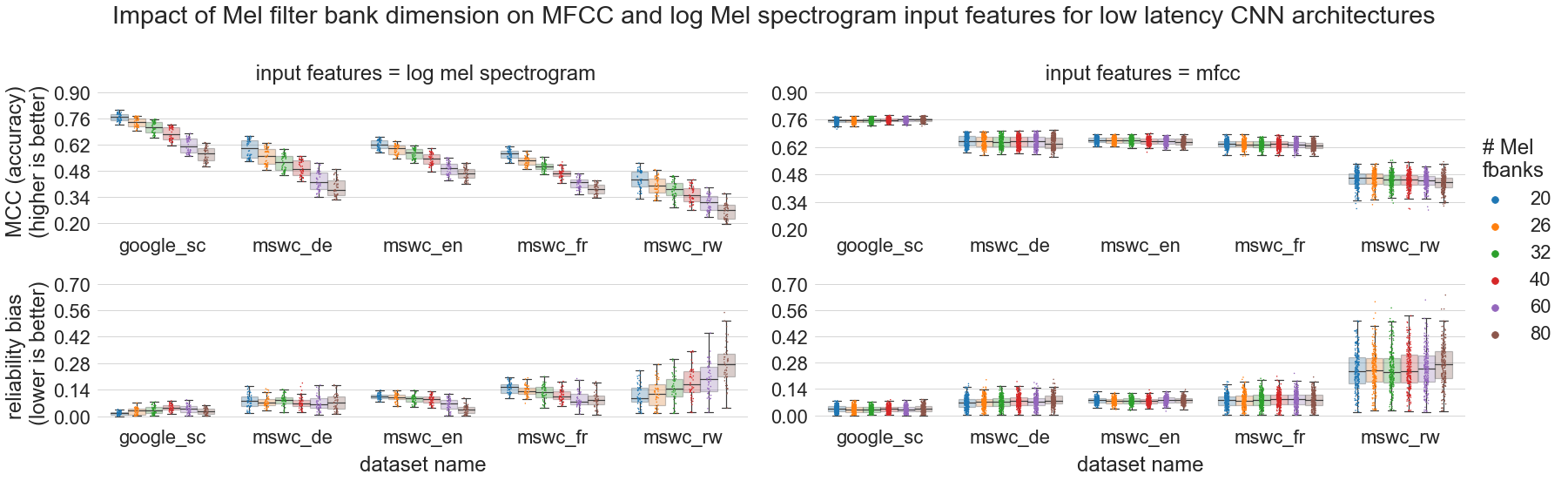}
    \caption{Effect of log Mel spectrogram dimensions (\# Mel fbanks) on accuracy and reliability bias, disaggregated by input feature type for \textbf{low latency CNN} architectures. The number of Mel filter banks clearly impacts models that directly use log Mel spectrograms as input features.}
\label{fig:dataset_arch_melbins_feature_type_llcnn}
\end{figure}

Figure~\ref{fig:dataset_arch_melbins_feature_type_llcnn} visualizes the impact of the number of Mel filter banks on accuracy and reliability bias for low latency CNN architectures. Figure~\ref{fig:dataset_arch_melbins_feature_type_cnn} in the Appendix visualizes results for CNN architectures, which show similar trends. It is clear that when models use log Mel spectrograms directly as input features (left column), the number of Mel filter bank dimensions has a critical impact on accuracy: MCC scores deteriorate rapidly as the number of Mel filter banks increases. The impact on reliability bias is more varied. For the mswc\_en and \_fr datasets the Mel filter bank dimensions pose a trade-off between accuracy (models with more filter banks are less accurate) and reliability bias (models with more filter banks are less biased). Models trained with the google\_sc and mswc\_de datasets show no clear trend. Only models trained with the mswc\_rw dataset have lower reliability bias for fewer Mel filter banks, thus allowing developers to choose Mel filter bank dimensions that increase accuracy while reducing bias. 

When used with MFCCs, log Mel spectrograms serve a purpose of dimensionality reduction. In contrast to log Mel spectrogram input features, MFCC features (right column) are robust to the number of Mel filter banks used across all datasets. Interestingly, when comparing the results of the low latency CNN models trained with MFCCs in this figure and the CNN models in Figure~\ref{fig:dataset_arch_melbins_feature_type_cnn} in the Appendix, the distributions of accuracy scores for the google\_sc and mswc\_en datasets have a smaller IQR for the light-weight architecture. This suggests that the dimensionality reducing effect of the spectrograms can be particularly advantageous for smaller model architectures. As fewer input dimensions reduce the computational overhead during training and inference, our results present an opportunity for on-device ML developers: MFCCs that use log Mel spectrograms with fewer filter banks (e.g. 20) can improve computational efficiency without compromising accuracy or reliability bias.  

\begin{figure}[hbt]
    \centering
    \includegraphics[width=0.8\textwidth]{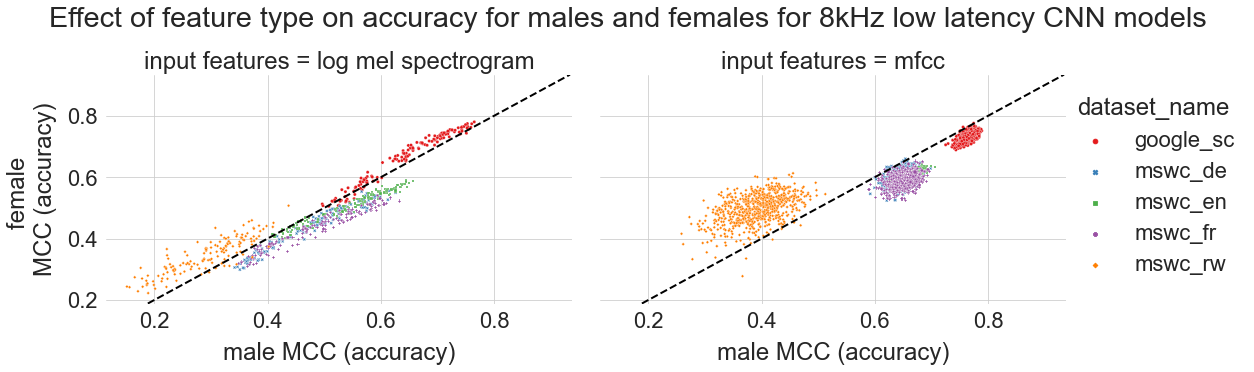}
    \caption{Accuracy scores for males (x-axis) and females (y-axis) for log Mel spectrogram (left) and MFCC (right) feature types for \textbf{8kHz low latency CNN} models.}
    \label{fig:feature_type_subgroups_llcnn8}
\end{figure}

To gain an appreciation of how pre-processing parameters affect the performance of KWS models for male and female subgroups, we show the impact of feature type on male and female subgroup accuracy in Figure~\ref{fig:feature_type_subgroups_llcnn8}. For mswc\_de, \_en and \_fr accuracy is always greater for males, irrespective of the feature type. For mswc\_rw the opposite holds true: accuracy is almost always greater for females, irrespective of the feature type. For the google\_sc dataset log Mel spectrograms generate models that have higher accuracy for females than for males, while MFCC features generate models with lower accuracy for females than males. For the MSWC datasets MFCC features clearly result in more accurate models for both subgroups while for the google\_sc dataset both feature types generate models with similar maximum accuracy. When training on this dataset the choice of feature type can thus be a source of reliability bias. 

As highlighted by a recent study in the speaker recognition domain~\cite{Liu2020comparative}, only limited feature extractors have been considered since the adoption of deep neural networks for speech processing tasks. Some prior studies have noted the limits of log Mel and MFCC based features, and have proposed alternatives. For example, per-channel energy normalization features have been proposed for robust keyword spotting~\cite{Wang2017trainable} and power-normalized cepstral coefficients for robust speech recognition~\cite{Kim2016power}. However, while these studies consider robustness in noisy and far-field environments, they do not consider bias in their analysis of robustness. Based on our findings we consider further characterisation of the effect of input features on reliability bias across a wider range of feature extractors an important area of future work.

\begin{mdframed}[style=greybox] \textbf{Key insights}: Feature type and dimensions impact KWS accuracy and reliability bias. Their effect is further influenced by the training dataset. In general, MFCC type features perform better than log Mel spectrograms. However, they can also increase reliability bias, prejudicing models against females and favouring males. For MFCC features, fewer dimensions (i.e. cepstral coefficients and Mel filter banks) can reduce computational demands with a negligible impact on accuracy and reliability bias. 
\end{mdframed}

\subsection{Design Choices during Model Optimization}
\label{ss:model_optimization_choices}

We focused our study of model optimization design choices on model compression and in particular model pruning. Pruning increases model sparsity, which reduces storage, memory and bandwidth requirements when downloading models to devices. We followed the experimental setup described in \S\ref{ss:training_details} to prune a subset of the most accurate and least biased models.

To analyze the impact of the pruning hyper-parameters, we performed factorial ANOVA tests to determine the effects of pruning hyper-parameters on change in reliability bias and change in accuracy due to pruning. The factorial ANOVA tests were designed following the same process as described for pre-processing parameters in the previous section. The first factorial ANOVA model (see Model~\ref{mod:anova4} in the Appendix) considers interactions between all the independent variables, including pruning hyper-parameters, dataset, architecture, sample rate, the baseline model accuracy and baseline model reliability bias. We continued to improve the factorial ANOVA models separately for change in accuracy and change in reliability bias by removing all non-significant interactions, and then including lower-level interactions. The final ANOVA models are included in the Appendix, with Models~\ref{mod:anova5} and~\ref{mod:anova6} capturing variables and interactions of pruning design choices on the change in MCC score and change in reliability bias respectively.

Tables~\ref{tab:pruning_hyperparameter_importance_mcc} and~\ref{tab:pruning_hyperparameter_importance_bias} show statistically significant interaction and main effects of the final factorial ANOVA models. For completeness we have included main effects even if they already contribute to an interaction. The final factorial ANOVA models are significant (change in MCC~(accuracy): $F(274)=555.74$, $p=0.0$, $R^2_{adj.}=0.9259$ and change in reliability bias: $F(148)=110.70$, $p=0.0$, $R^2_{adj.}=0.5717$). We again point the reader to Table~\ref{tab:fcrit_at_p} for reference of the critical F statistics at p-values less than 0.01 and 0.05. Based on the F statistics, we reject the null hypothesis that KWS model accuracy and reliability bias are unaffected by pruning hyper-parameters and their interactions during model optimization. As with the statistical analysis of the pre-processing parameters, we found that the $R^2$ values of the change in accuracy ANOVA model ($R^2=0.9276$, $R^2_{adjusted}=0.9259$) indicate that this model captures the effects better than the change in reliability bias ANOVA model ($R^2=0.5769$, $R^2_{adjusted}=0.5717$). In the latter model a portion of the variance in the dependent variable remains unaccounted for. We now examine the impact of the pruning interaction effects in greater detail. Throughout the analysis we use the terms \emph{change in} and \emph{delta} interchangeably.

\setlength{\tabcolsep}{3pt}
\setlength{\textfloatsep}{0.5cm}
\renewcommand{\arraystretch}{1.1}
\begin{table}[hbt]
\centering
\resizebox{\linewidth}{!}{
\begin{tabular}{lcccc}
\toprule
\textbf{Factorial ANOVA main and interaction effects} & \textbf{SS} & \textbf{df} & \textbf{F} & \textbf{p(\textless{}0.05)} \\
\midrule
pruning schedule & 6.2660 & 1 & 2.9467E+03 & 0.0E+00 \\
reliability bias baseline model & 0.7614 & 1 & 3.5805E+02 & 1.1E-78 \\
dataset & 6.4134 & 4 & 7.5402E+02 & 0.0E+00 \\
pruning learning rate & 89.6026 & 2 & 2.1069E+04 & 0.0E+00 \\
final sparsity & 143.5794 & 5 & 1.3504E+04 & 0.0E+00 \\ 
dataset * pruning schedule * final sparsity & 0.2113 & 20 & 4.9694E+00 & 1.9E-12 \\
sample rate * pruning learning rate * final sparsity & 0.2759 & 10 & 1.2974E+01 & 7.2E-23 \\
dataset * pruning learning rate * pruning schedule & 0.1467 & 8 & 8.6237E+00 & 8.5E-12 \\
dataset * pruning learning rate * final sparsity & 2.7539 & 40 & 3.2377E+01 & 3.7E-232 \\
model architecture * pruning learning rate * pruning schedule * final sparsity & 0.2685 & 10 & 1.2625E+01 & 3.6E-22 \\
dataset * model architecture * sample rate * final sparsity & 0.1929 & 20 & 4.5357E+00 & 6.2E-11 \\
Residual & 25.2897 & 11893 & - & - \\
\midrule
\multicolumn{1}{r}{\textbf{Model}} & - & 274 & 5.5574E+02 & 0.0E+00 \\
\multicolumn{1}{r}{\textbf{$R^{2}$}} & 0.9276 &  &  &  \\
\multicolumn{1}{r}{\textbf{Adjusted $R^{2}$}} & 0.9259 &  &  & \\ \hhline{=====}
\end{tabular}}
\smallskip
\caption{Significant main and interaction effects of pruning hyper-parameters on \textbf{change in MCC (accuracy)}. SS=sum of squares, df=degrees of freedom}
\label{tab:pruning_hyperparameter_importance_mcc}
\end{table}

\setlength{\tabcolsep}{5pt}
\setlength{\textfloatsep}{0.5cm}
\renewcommand{\arraystretch}{1.1}
\begin{table}[hbt]
\centering
\resizebox{\linewidth}{!}{
\begin{tabular}{lcccc}
\toprule
\textbf{Factorial ANOVA main and interaction effects} & \textbf{SS} & \textbf{df} & \textbf{F} & \textbf{p(\textless{}0.05)} \\
\midrule
model architecture & 2.0088 & 1 & 2.2874E+02 & 3.3E-51 \\
pruning learning rate & 15.0129 & 2 & 8.5475E+02 & 0.0E+00 \\
final sparsity & 23.8760 & 5 & 5.4374E+02 & 0.0E+00 \\
reliability bias baseline model & 8.5336 & 1 & 9.7171E+02 & 3.3E-205 \\
dataset & 22.9815 & 4 & 6.5421E+02 & 0.0E+00 \\
pruning schedule * final sparsity & 1.0450 & 5 & 2.3798E+01 & 6.8E-24 \\
model architecture * final sparsity & 2.8381 & 5 & 6.4633E+01 & 8.5E-67 \\
dataset * pruning learning rate * final sparsity & 10.8173 & 40 & 3.0794E+01 & 3.2E-220 \\
dataset * model architecture * sample rate * pruning learning rate & 1.1197 & 8 & 15.937276 & 1.3E-23 \\
Residual & 105.508229 & 12014 & - & - \\
\midrule
\multicolumn{1}{r}{\textbf{Model}} & - & 148 & 1.1070E+02 & 0.0E+00 \\
\multicolumn{1}{r}{\textbf{$R^{2}$}} & \multicolumn{1}{l}{0.5769} &  &  &  \\
\multicolumn{1}{r}{\textbf{Adjusted $R^{2}$}} & \multicolumn{1}{l}{0.5717} &  &  & \\ \hhline{=====}
\end{tabular}}
\smallskip
\caption{Significant main and interaction effects of pruning hyper-parameters on \textbf{change in reliability bias}. SS=sum of squares, df=degrees of freedom}
\label{tab:pruning_hyperparameter_importance_bias}
\end{table}

\subsubsection{Impact of Pruning Hyper-Parameters}
\label{ss:impact_of_pruning_hyperparameters}

\begin{figure}[hbt]
    \centering
    \includegraphics[width=\textwidth]{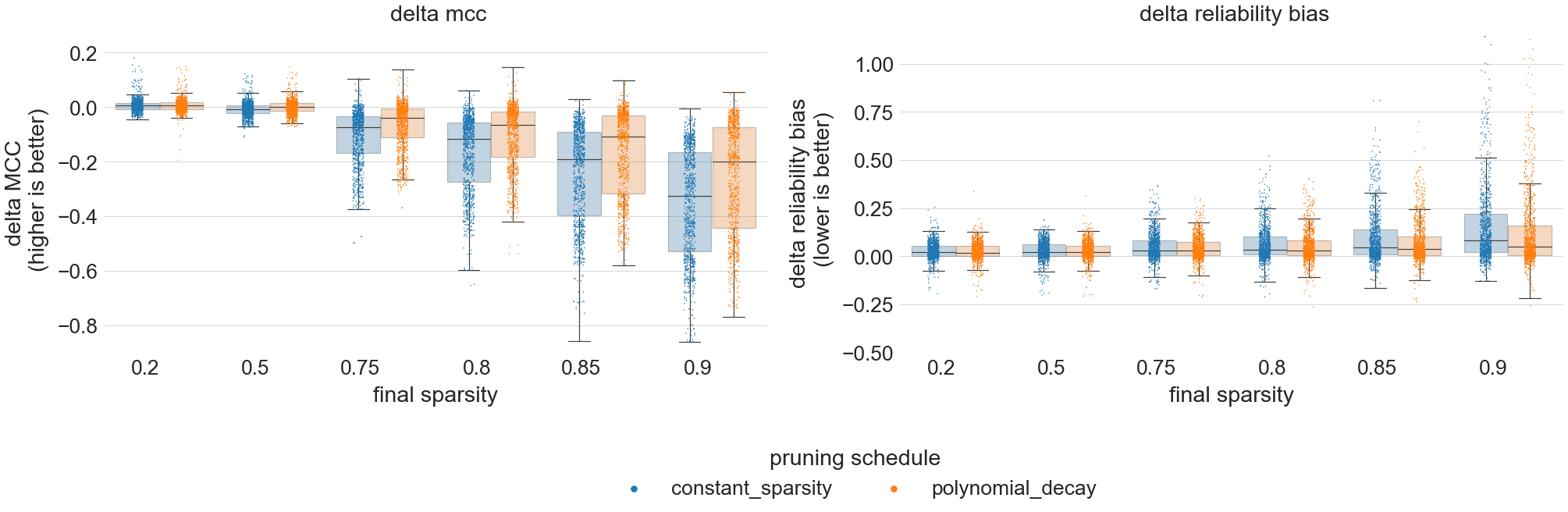}
    \caption{Interaction effect of final sparsity and pruning schedule on \textbf{change in MCC (accuracy)} (left) and \textbf{change in reliability bias} (right). The \emph{polynomial decay} (orange) pruning schedule results in higher median change in MCC scores and lower median change in reliability bias. Polynomial decay also results in smaller IQR of reliability bias. These effects become more significant as final sparsity increases.}
    \label{fig:pruninghps_sparsity-schedule}
\end{figure}

The interaction effect between final sparsity and pruning schedule is visualized in Figure~\ref{fig:pruninghps_sparsity-schedule}. This interaction significantly affects change in reliability bias due to pruning (as per Table~\ref{tab:pruning_hyperparameter_importance_bias}). The interaction between final sparsity, pruning schedule and dataset also have a significant effect on change in accuracy (as per Table~\ref{tab:pruning_hyperparameter_importance_mcc}). The figure highlights several interesting observations. When final sparsities are low (i.e. 0.2 and 0.5), the median delta MCC and delta reliability bias are close to zero. Furthermore, the delta MCC and delta reliability bias IQR of models pruned to these sparsities are small. This indicates that these pruned models have low variability in accuracy and reliability bias and that scores lie close to those of the baseline models. However, as the final sparsity increases, the median delta MCC becomes more negative (implying lower accuracy due to pruning) and the IQR increases (indicating greater variability in accuracy due to pruning). Likewise, the median delta reliability bias and the IQR increase, indicating that models become more biased and that reliability bias scores become more variable. For all sparsities there are some models that have a positive change in MCC, thus becoming more accurate, and a negative change in reliability bias, thus becoming less biased, due to pruning. The polynomial decay pruning schedule results in higher median delta MCC scores and lower median delta reliability bias. Polynomial decay also results in smaller IQR of delta reliability bias. These effects become more apparent as final sparsity increases. For developers polynomial decay is thus a more robust pruning schedule to choose.

\begin{figure}[hbt]
    \centering
    \includegraphics[width=\textwidth]{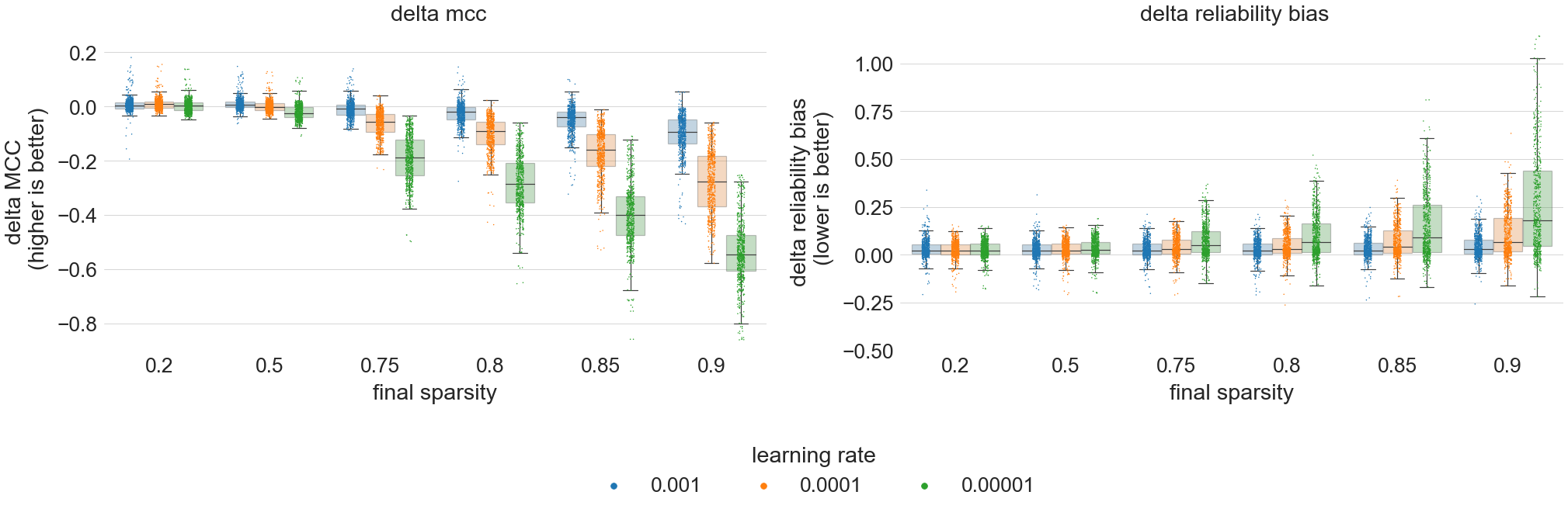}
    \caption{Interaction effect of final sparsity and pruning learning rate on \textbf{change in accuracy} (left) and \textbf{change in reliability bias} (right). At final sparsities above 0.5 smaller learning rates significantly reduce MCC scores and increase reliability bias.}
    \label{fig:pruninghps_sparsity-lr}
\end{figure}

Figure~\ref{fig:pruninghps_sparsity-lr} visualizes the interaction effect of final sparsity and the pruning learning rate. As shown in Table~\ref{tab:pruning_hyperparameter_importance_bias} this interaction significantly affects the change in reliability bias due to pruning. Final sparsity and pruning learning rate also have a significant effect on change in accuracy through interactions with sample rate and with dataset (see Table~\ref{tab:pruning_hyperparameter_importance_mcc}). At a low final sparsity of 0.2 the learning rate has no impact on the accuracy and bias of pruned models. As the sparsity increases, this changes dramatically. The smaller the learning rate, the lower the median delta MCC and the larger its IQR. A lower delta MCC results in a greater accuracy drop due to pruning. Similarly, the smaller the learning rate, the higher the median delta reliability bias and the larger its IQR. A higher delta reliability bias results in increase in reliability bias due to pruning. At 90\% sparsity the median MCC score of models pruned with a learning rate of 0.00001 reduces by more than 0.5 (maximum value of the MCC metric is 1). This means that the accuracy of pruned models with 90\% sparsity is less than half that of baseline models. At the same final sparsity and learning rate the median delta reliability bias increases by 0.18, indicating that substantial performance discrepancies exist between the male and female subgroups. 

A possible explanation for our results is that the learning rate optimizes the discovery of structure in the training data to favour one subgroup over the other. This intuition aligns with recent empirical work that shows that top performing deep neural networks can have very similar accuracy, but large variance in other performance aspects such as inference latency due to hyper-parameter tuning~\cite{Liao2022empirical}. Similarly, a recent study on fixed-seed training of deep learning systems shows high variance in fairness measures if experiments consist of a single run with a fixed seed~\cite{Qian2021are}. Based on our results, developers can choose a larger pruning learning rate, like 0.001, during model optimization to reduce the likelihood of unintended bias and unexpected accuracy degradation, especially when pruning models to high sparsities. While this rule-of-thumb is useful given our current knowledge, further research is needed to fully characterize the impact of the pruning learning rate on model performance and bias. We thus suggest that developers empirically validate and optimize the learning rate during pruning.    

\begin{figure}[hbt]
    \centering
    \includegraphics[width=\textwidth]{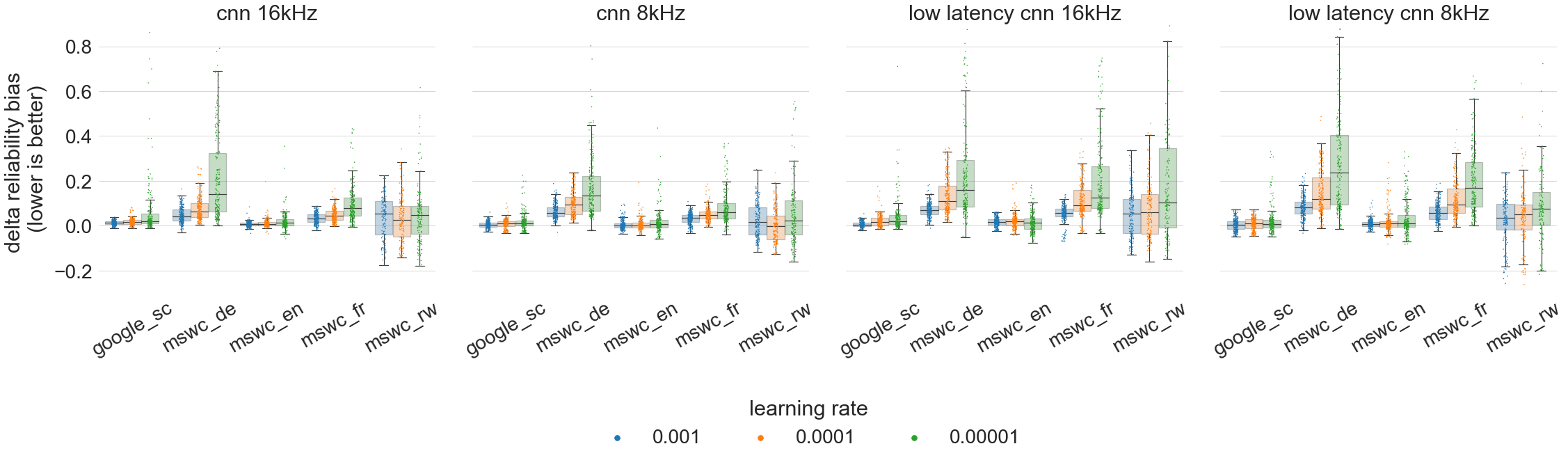}
    \caption{Interaction effect between dataset, architecture, sample rate and pruning learning rate on \textbf{change in reliability bias}. Across datasets the general trend indicates that the smaller the learning rate, the more biased models become. The median change in reliability bias of the google\_sc and mswc\_en datasets is less affected by the pruning learning rate than that of the remaining datasets.}
    \label{fig:pruninghps_dataset-lr}
\end{figure}

To conclude our detailed analysis of interaction effects arising during model pruning, we examine how the interactions between dataset, architecture, sample rate and pruning learning rate affect change in reliability bias in Figure~\ref{fig:pruninghps_dataset-lr}. Across datasets, architectures and sample rates we observe the general trend which we have already identified in the previous figure: the smaller the learning rate, the more biased models become. Careful examination of the results across architectures and sample rates also reveals trends similar to those we observed with pre-processing parameters: the increase in reliability bias due to pruning is greater for the light-weight low latency CNN architecture and the lower sample rate of 8kHz, as indicated by higher medians and larger IQRs. This trend is stronger at smaller learning rates. As with the pre-processing parameters, the mswc\_rw dataset presents an exception to this observation. For this dataset median delta reliability bias across learning rates shows no clear trend, while the IQRs are always large when compared to the IQRs of the other datasets. 

Figure~\ref{fig:pruninghps_dataset-lr} reveals a further insight when comparing results across datasets. The reliability bias of models trained on the google\_sc and mswc\_en datasets is less affected by the pruning learning rate than what is the case for models trained on the remaining datasets. These two English language datasets have low median delta reliability bias values and a small IQR. A likely contributor to these results is that English is the best resourced language, with the largest available quantity of data. The English datasets in our study thus include more utterances per keyword, more unique speakers per keyword and better representation of speakers and utterances across keywords in the validation and test sets. Data quantity and representation, however, may not explain the entire effect. The mswc\_de and mswc\_fr datasets have very similar statistics across the keywords, genders and dataset splits, with the mswc\_de dataset being 13\% larger than the mswc\_fr dataset. Yet, the change in reliability bias for mswc\_de models is larger and more variable than that of mswc\_fr models. Further research is needed to understand the source of this variability. For developers, our results highlight that training, validation and test datasets need to be large enough and representative across groups of users to ensure robust results and avoid bias. Considering that German and French are, after English, two of the best resourced languages in the Mozilla Common Voice corpus\footnote{https://commonvoice.mozilla.org/en/languages}, this may mean in practice that developers need to collect context and application specific datasets to evaluate bias.

\begin{mdframed}[style=greybox] \textbf{Key insights}: Polynomial decay is a more robust pruning schedule than constant sparsity, and a larger pruning learning rate, like 0.001, reduces the likelihood of unintended bias and unexpected accuracy degradation. These design choices are particularly important when pruning models to sparsities greater than 50\%, beyond which accuracy and reliability bias can deteriorate dramatically. The increase in reliability bias due to pruning is greater for smaller architectures and at lower sample rates. This trend is stronger at smaller learning rates. Training, validation and test datasets need to be large enough and representative across groups of users to ensure robust results and avoid bias.
\end{mdframed}

\subsection{Summary of Results}
We conducted empirical experiments for an audio KWS task to investigate the impact of a comprehensive set of design choices on accuracy and reliability bias during model training and optimization. During model training we investigated design choices related to the data sample rate and pre-processing parameters. We also considered how models trained with a light-weight architecture are affected by the sample rate and pre-processing parameters. Analyzing the results of 17280 experiments on 5 datasets showed that median and interquartile range (IQR) of reliability bias tend to be greater at lower sample rates and for light-weight architectures. Whether reliability bias favours or is prejudiced against a group of people was strongly influenced by the training dataset. Overall, male speakers were favoured by models. Model accuracy was lower at lower sample rates and for light-weight architectures. An exception to the overall trends were models trained on the mswc\_rw dataset, which had considerably lower accuracy and favoured female speakers.

With regards to pre-processing parameters, we found feature type and dimensions to impact KWS accuracy and reliability bias. These effects were further influenced by the training dataset. In general, MFCC type features performed better than log Mel spectrograms. However, in some instances they also increased reliability bias, prejudicing models against females and favouring males. For MFCC features, reducing the feature dimensions by using fewer Mel filter banks and fewer cepstral coefficients had a negligible impact on accuracy and reliability bias. This presents an opportunity to reduce computational demands for on-device ML applications.

During model optimization we investigated design choices related model compression, and specifically pruning. Analyzing the results of 12168 experiments on 5 datasets, we found polynomial decay to be a more robust pruning schedule than constant sparsity. The smaller the pruning learning rate, the more pruning increased reliability bias and decreased accuracy of baseline models. We found that a larger pruning learning rate, like 0.001, reduced the change in reliability bias and accuracy. These design choices were particularly important when models were pruned to sparsities greater than 50\%. Beyond this, accuracy and reliability bias deteriorated dramatically. As with pre-processing parameters, the increase in reliability bias due to pruning was greater for light-weight architectures and at lower sample rates. This trend was stronger at smaller learning rates. Finally, we found that pruning results varied across datasets, with English language datasets showing a smaller increase in reliability bias due to pruning than other languages. One take-away from this is that training, validation and test datasets need to be large enough and representative across groups of users to ensure robust results and avoid bias.

%% file: sections/7_discussion.tex
\section{Strategies to Mitigate Reliability Bias}
\label{s:mitigation}

In the previous section we presented empirical results and an analysis of the impact of design choices on accuracy and reliability bias for an audio KWS task. Taking the insights gained through the study into consideration, we now offer low effort strategies for mitigating reliability bias. We first consider strategies for model selection and then discuss approaches for supporting design choices with targeted experimentation.

\subsection{Model Selection}
\label{ss:model_selection}
In the decision map presented in Figure~\ref{fig:conceptual_model_design_choices} model selection can occur after training a new model, downloading pre-trained models or optimizing a model. We consider model selection strategies that account for accuracy and reliability bias after model training and after model optimization. Rather than considering reliability bias and accuracy as a trade-off, we seek approaches that enable engineers to navigate multi-objective search scenarios where high accuracy and low reliability bias are desired. In contrast to multi-objective criteria that have been proposed during model training~\cite{Padh2021addressing, Liu2022accuracy}, here we focus on multi-objective model selection as a post-processing intervention.

\subsubsection{Model Selection After Training}

In \S\ref{ss:model_training_choices} we explored in depth how model training design choices impact reliability bias and accuracy. While our analysis presented important insights of trends that exist across datasets and architectures, we also found that models exist that are accurate and that perform equally well for male and female subgroups. A visual appreciation for this can be gained from Figure~\ref{fig:subgroup_performance_accuracy}. Across datasets, architectures and sample rates there are models that lie on or close to the diagonal on which male and female accuracy is equal. This suggests that pre-processing parameters may exist that produce models with high accuracy and low bias. However, these models do not necessarily have the highest accuracy score. We thus considered search criteria for selecting models based on accuracy and reliability bias. Listed below are three criteria we used to select model $d$ from $n$ trained models $D$ by optimizing for:

\begin{enumerate}
    \item high accuracy: select $d$ if $MCC_d=max(MCC_1, ..., MCC_n)$ for $n$ in $D$
    \item low bias: select $d$ if $reliability\ bias_d=min(reliability\ bias_1, ..., reliability\ bias_n)$ for $n$ in $D$
    \item low bias + high accuracy: select $d$ if  $MCC_d>=0.985*max(MCC_1, ..., MCC_n)$ for $n$ in $D$ \textbf{and}\\ $reliability\ bias_d=min(reliability\ bias_1, ..., reliability\ bias_k)$ for $k$ where $MCC_k>=0.985*max(MCC_1, ..., MCC_n)$ for $n$ in $D$
\end{enumerate}

We consider the high accuracy criteria as a baseline, as this is the typical strategy followed by engineers that do not consider bias. The low bias criteria presents the opposite scenario, where only reliability bias informs model selection. Finally, the low bias + high accuracy criteria considers accuracy as a satisficing metric while minimizing reliability bias. This criteria selects the model with the lowest reliability bias, provided that it has an accuracy score within a 1.5\% threshold of the maximum accuracy. A reasonable threshold value should be selected in accordance with the application requirements. The multi-objective approach allows us to explore alternative models for deployment.

Table \ref{tab:model_selection} shows the MCC (accuracy) score and reliability bias for the best models trained on the google\_sc dataset, selected according to the three criteria. We find that the low bias + high accuracy criteria selects models with a low reliability bias across architectures, while retaining an MCC score close to the high accuracy criteria. For the CNN architectures, this criteria reduces reliability bias by 15.7 and 1.7 fold for models trained with 16kHz and 8kHz sample rates respectively. For the 8kHz low latency CNN model, reliability bias is reduced 22.3 fold. For the 16kHz low latency CNN architecture the model with the highest accuracy also has the lowest reliability bias and thus experiences no reduction. By comparison, models selected using only low bias as selection criteria have a very low reliability bias. However, this comes at the cost of an accuracy drop between 3.2\% and 6.1\%, which is considerably greater than the desired 1.5\% threshold and will degrade performance for both subgroups.

\begin{table}[hbt]
\centering
\small
\begin{tabular}{l|lcccc}
\thead{model selection\\criteria} & \thead{\\metric} & \thead{16kHz\\CNN} & \thead{8kHz\\CNN} & \thead{16kHz\\low latency CNN} & \thead{8kHz\\low latency CNN} \\\midrule
high accuracy & MCC score & 0.877  & 0.868  & 0.804  & 0.778  \\
                & reliability bias & 1.2e-2 & 9.8e-3 & 6.6e-4 & 4.1e-2 \\ \midrule
low bias & MCC score & 0.849  & 0.815  & 0.762  & 0.740  \\
                & reliability bias & 1.8e-4 & 1.9e-4 & 1.2e-4 & 1.6e-4 \\ \midrule
low bias + high accuracy & MCC score & 0.872  & 0.861  & 0.804  & 0.775  \\
                & reliability bias & 7.7e-4 & 5.9e-3 & 6.6e-4 & 1.8e-3
\end{tabular}
\smallskip
\caption{Table of MCC (accuracy) scores and reliability bias for models trained on the google\_sc dataset and selected for top performance based on three criteria: 1) high accuracy 2) low bias 3) low bias + high accuracy. Comparison across metrics shows that the low bias + high accuracy criteria, which accepts a marginal drop in accuracy of up to 1.5\%, selects models with considerably lower bias than the high accuracy strategy.}
\label{tab:model_selection}
\end{table}

Instead of selecting maximum or minimum values, the selection criteria can be modified to select the $m$ best models under that criteria. We followed this approach choosing $m=3$ best models for a dataset, model architecture and sample rate triplet to select baseline models for the pruning experiments. The low bias + high accuracy criteria did not return valid models for all triplets, which is the reason for the unequal number of baseline models in Table~\ref{tab:pruning_experiments}. Next we consider how these selection criteria hold up after pruning.

\subsubsection{Model Selection After Pruning}
\begin{figure}[hbt]
        \centering
        \includegraphics[width=\textwidth]{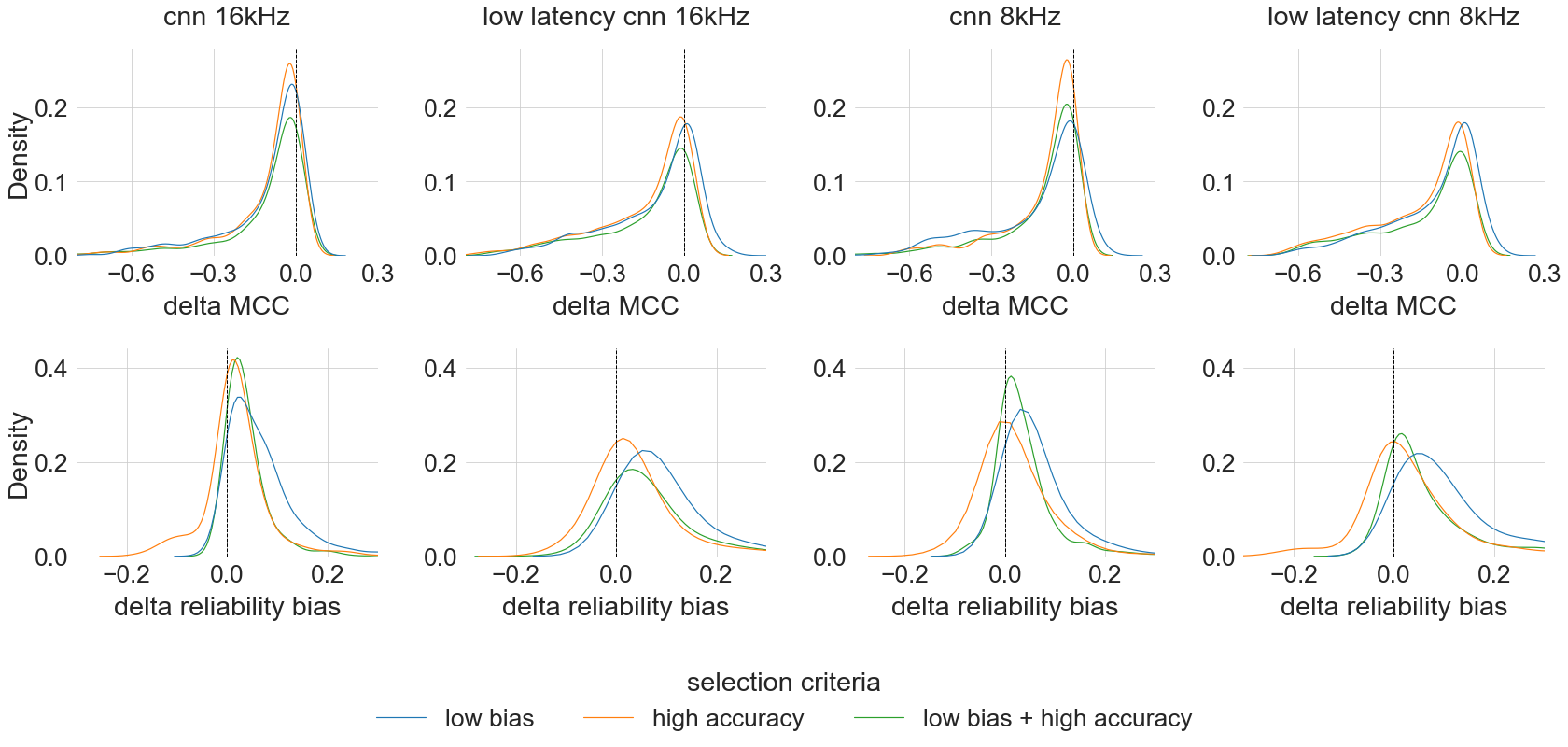}
        \caption{\textbf{Change in MCC (accuracy)} score and \textbf{change in reliability bias} of models after pruning. Models were selected based on three selection criteria: high accuracy (orange), low bias (blue), and high accuracy + low bias (green). Delta MCC is better when greater, delta reliability bias is better when smaller. The selection criteria has no evident effect on delta MCC, but does affect delta reliability bias.}
        \label{fig:pruninghps_selection_criteria_delta}
\end{figure}

For the pruning experiments in \S\ref{ss:model_optimization_choices} we selected the top 3 models per architecture and sample rate for each model selection criteria. We now consider the post-pruning performance of models selected with different criteria. In Figure~\ref{fig:pruninghps_selection_criteria_delta} we show the delta (i.e. change in) MCC score (top) and delta reliability bias (bottom) after pruning, for models selected under the three criteria after training. In the density distributions in the figure accuracy increases in the direction of positive change, meaning that delta MCC distributions that peak to the right of zero are desirable. Conversely, reliability bias decreases in the direction of negative change, meaning that delta reliability bias distributions with peaks to the left of zero are desirable. The delta MCC distributions peak just left of zero (CNN architectures) or on zero (low latency CNN architectures), indicating that the majority of models with these architectures experience a decline in accuracy. The shape of the distributions is similar for different selection criteria under the same architecture and sample rate, with the accuracy of low bias models (which have lower baseline accuracy) increasing slightly more after pruning.

The shapes and peaks of the delta reliability bias distributions vary across model selection criteria. This indicates that the model selection criteria impact reliability bias after pruning. A further confirmation of this is presented in the statistical analysis in Table~\ref{tab:pruning_hyperparameter_importance_bias}, where the reliability bias of the baseline model has a statistically significant effect on delta reliability bias due to pruning. Analyzing the distributions, we can see that the distributions of models selected for low bias (blue) lie furthest to the right. This means that the reliability bias of these models increases the most after pruning. This is not surprising, as the lower bound of the reliability bias measure is zero and models with low bias are very sensitive to small changes in reliability bias. However, this can also indicate that models selected for low bias may loose some of their good bias properties during pruning. The distributions of models selected for high accuracy (orange) lie furthest to the left. These models typically started out with higher reliability bias after training, which makes them less sensitive to changes in reliability bias and thus better able to retain their reliability bias scores. 

In Figure~\ref{fig:pruninghps_selection_criteria_metric} in the Appendix we visualize the distribution of MCC scores and reliability bias for the selection criteria. Right of the peak (i.e. in the higher accuracy range), the MCC score distributions for the high accuracy criteria and the low bias + high accuracy criteria lie very close to each other. After pruning models selected for high accuracy and for low bias + high accuracy thus have similar MCC scores. For reliability bias the distribution of the low bias + high accuracy criteria lies between the low bias and high accuracy distributions. The low bias + high accuracy criteria thus results in models with lower bias after pruning than the high accuracy criteria. Overall, this makes the low bias + high accuracy criteria a good choice to select a range of models for pruning.

Finally we reapply the same model selection criteria that we previously applied after training, \emph{after pruning}. Table~\ref{tab:pruning_performance_across_sparsities} shows the mean and standard deviation of accuracy and reliability bias across sparsities for the three selection criteria for \emph{pruned} models trained on google\_sc. We observe that mean reliability bias can be improved by an order of magnitude by choosing the low bias + high accuracy criteria rather than the high accuracy criteria. Models selected with the low bias criteria suffer a large drop in accuracy. While the low bias criteria offers lower reliability bias than the low bias + high accuracy criteria,  the latter already has a low mean and variance in reliability bias, making additional reductions less impactful. For all models the variance of metrics across sparsities is relatively low, which is supported by our earlier observation that models trained on the google\_sc dataset are less affected by pruning hyper-parameters (see Figure~\ref{fig:pruninghps_dataset-lr}). Across all datasets the low bias + high accuracy criteria selects models with similar accuracy and lower reliability bias than the high accuracy criteria. This outcome is not surprising, as the purpose of a multi-objective criterion is precisely to satisfy multiple objectives. The value of our analysis lies in empirically validating the obvious rather than in finding surprise: engineers can reduce bias in audio KWS with little effort by applying a multi-objective criterion during model selection, choosing models that satisfy an accuracy condition while minimizing bias.   

\setlength{\tabcolsep}{4pt}
\setlength{\textfloatsep}{0.5cm}
\begin{table}[hbt]
\centering
\resizebox{\linewidth}{!}{
\begin{tabular}{c|llll|llll|llll}
\textbf{criteria} & \multicolumn{4}{c|}{\textbf{high accuracy}} & \multicolumn{4}{c|}{\textbf{low bias + high accuracy}} & \multicolumn{4}{c}{\textbf{low bias}} \\
\textbf{metric} & \multicolumn{2}{c}{\textbf{MCC score}} & \multicolumn{2}{c|}{\textbf{\emph{reliability bias}}} & \multicolumn{2}{c}{\textbf{MCC score}} & \multicolumn{2}{c|}{\textbf{\emph{reliability bias}}} & \multicolumn{2}{c}{\textbf{MCC score}} & \multicolumn{2}{c}{\textbf{\emph{reliability bias}}} \\
\multicolumn{1}{l|}{} & \multicolumn{1}{c}{\textbf{mean}} & \multicolumn{1}{c|}{\textbf{var}} & \multicolumn{1}{c}{\textbf{mean}} & \multicolumn{1}{c|}{\textbf{var}} & \multicolumn{1}{c}{\textbf{mean}} & \multicolumn{1}{c|}{\textbf{var}} & \multicolumn{1}{c}{\textbf{mean}} & \multicolumn{1}{c|}{\textbf{var}} & \multicolumn{1}{c}{\textbf{mean}} & \multicolumn{1}{c|}{\textbf{var}} & \multicolumn{1}{c}{\textbf{mean}} & \multicolumn{1}{c}{\textbf{var}}\\ \midrule
\textbf{16kHz CNN} & 0.885 & \multicolumn{1}{l|}{1.2e-02} & 1.4e-02 & \multicolumn{1}{l|}{9.6e-03} & 0.879 & \multicolumn{1}{l|}{1.1e-02} & 4.0e-03 & \multicolumn{1}{l|}{5.5e-03} & 0.823 & \multicolumn{1}{l|}{5.3e-02} & 6.5e-04 & 4.0e-04 \\ {\medskip}
\textbf{8kHz CNN} & 0.876 & \multicolumn{1}{l|}{9.2e-03} & 6.5e-03 & \multicolumn{1}{l|}{1.8e-03} & 0.870 & \multicolumn{1}{l|}{8.9e-03} & 1.1e-03 & \multicolumn{1}{l|}{6.7e-04} & 0.851 & \multicolumn{1}{l|}{1.9e-02} & 2.9e-04 & 2.6e-04 \\ 
\textbf{16kHz llCNN} & 0.808 & \multicolumn{1}{l|}{1.8e-02} & 8.9e-03 & \multicolumn{1}{l|}{7.0e-03} & 0.804 & \multicolumn{1}{l|}{1.7e-02} & 1.4e-03 & \multicolumn{1}{l|}{1.1e-03} & 0.772 & \multicolumn{1}{l|}{2.3e-02} & 4.9e-04 & 4.1e-04 \\ {\smallskip}
\textbf{8kHz llCNN} & 0.785 & \multicolumn{1}{l|}{2.5e-02} & 1.0e-02 & \multicolumn{1}{l|}{7.6e-03} & 0.781 & \multicolumn{1}{l|}{2.4e-02} & 1.8e-03 & \multicolumn{1}{l|}{2.1e-03} & 0.761 & \multicolumn{1}{l|}{3.5e-02} & 4.9e-04 & 4.9e-04
\end{tabular}} 
\smallskip
\caption{Mean and variance of MCC scores and reliability bias across pruning sparsities (0.2, 0.5, 0.75, 0.8, 0.85, 0.9) for the three model selection criteria. Models have been trained on the google\_sc dataset. Models with lower bias can be selected for all sparsities at an accuracy cost of less than 1.5\%.}
\label{tab:pruning_performance_across_sparsities}
\end{table}

\subsubsection{Summary of Model Selection Strategy} 
Engineers should use a multi-objective criterion that considers accuracy and reliability bias to select models that have high accuracy and low bias after training or after pruning. We propose that engineers set a tolerance that controls the drop in accuracy from the maximum value, thus using accuracy as a satisficing metric while minimizing reliability bias. The tolerance value should be determined from application requirements. If model training is followed by pruning, a small number of top models should be selected for pruning using high accuracy and low bias + high accuracy strategies.

\subsection{Supporting Design Decisions with Targeted Experimentation} 

In \S\ref{s:ondevice} we presented a map of design choices arising in the on-device ML workflow. We then showed empirically that these design choices can lead to disparate performance of audio KWS models for males and females. Our analysis in \S\ref{s:results} demonstrates that even when engineers make reasonable decisions about training and optimization parameters (see Table~\ref{tab:experiment_design}) their choices can lead to models with widely different accuracy and bias properties. Especially when training light-weight architectures or processing data at low sample rates, systematic experimentation is a necessary strategy to support design decisions and mitigate bias. We have demonstrated that iterating over pre-processing parameters during training, and pruning hyper-parameters during model compression can help engineers train models with high accuracy and low bias. However, experimentation comes at a cost: each iteration requires computational resources, takes time and consumes energy. Where a single audio KWS model takes only a couple of minutes to train, we trained 17280 models, pruned 12168 models and ran our experiments for several days. This is a costly undertaking.

\begin{table}[hbt]
\resizebox{\linewidth}{!}{
    \centering
    \begin{tabular}{lllp{0.32\linewidth}}
        \textbf{Design action} & \textbf{Design choice} & \textbf{Choice variable (unit)} & \textbf{Variable values} \\ \midrule
        Train new model & input features | sample rate & sample rate (kHz)  & \textit{determined by application} \\ \noalign{\medskip}
        Train new model & input features | pre-processing & feature type & MFCC \\
        Train new model & input features | pre-processing & \# Mel filter banks & 20, 32 \\
        Train new model & input features | pre-processing & \# MFCCs & 10, 11 \\
        Train new model & input features | pre-processing & frame length (ms) & 20, 25, 30, 40 \\
        Train new model & input features | pre-processing & frame step (\% frame length) & 40, 50, 60\\
        Train new model & input features | pre-processing & window function & Hamming\\ \noalign{\medskip}
        Optimize model & light-weight architecture & model architecture & \textit{determined by application} \\ \noalign{\medskip}
        Optimize model & model compression | pruning & final sparsity (\%) & \textit{determined by application} \\
        Optimize model & model compression | pruning & pruning frequency & 10, 100\\
        Optimize model & model compression | pruning & pruning schedule & polynomial decay\\
        Optimize model & model compression | pruning & pruning learning rate & 1e-4, 1e-5 for sparsities < 50\%; 1e-3 for sparsities > 50\%\\ \noalign{\medskip}
        Model selection & selection strategy & criteria & high accuracy, low bias + high accuracy\\ 
        Model selection & selection strategy & \# best models & 3 \\ \noalign{\medskip}
    \end{tabular}}
    \caption{Recommended design choice variables and values for audio KWS to mitigate bias while reducing resource consumption during experimentation}
    \label{tab:recommended_design_choices}
\end{table}

Rather than replicating our approach, engineers should take the resource footprint and cost of model training into account, and target their experiments to iterate over values that are likely to yield high accuracy, low bias models. To this end we propose revised design choice variable values based on the insights we gained through our experiments in Table~\ref{tab:recommended_design_choices}. Given these reduced options, engineers only need to train 48 models per sample rate and architecture, and run at most 24 pruning experiments for low sparsities (12 experiments for higher sparsities of more than 50\%). This targeted approach to experimentation is thus a feasible strategy for engineers to use data-driven decision making to mitigate bias in on-device ML workflows.

\section{Discussion}
\label{s:discussion}

We now take a higher level perspective to reflect on the overarching implications of our work on bias in on-device ML. We first discuss reliability bias as a source of unfairness and discrimination in on-device ML and then reflect on limitations of the study.

\subsection{Reliability Bias as a Source of Unfairness and Discrimination in On-device ML}

On-device ML applications are becoming increasingly prevalent in our day-to-day lives, as consumers' privacy concerns and large volumes of sensor data are motivating a shift to run deep neural networks on devices rather than the cloud. Despite the prevalence of on-device ML applications, known bias challenges in ML systems, and the material consequences of system failure, bias in on-device ML is understudied. In this paper we set out to study sources of bias in on-device ML that have not been considered in the domain, and that are overlooked in current research on ML fairness. 

When interacting with services that make use of on-device ML, users are justified to expect reliable performance, irrespective of their demographic, social or economic attributes. We defined reliability bias as systematic device failures due to on-device ML performance disparities across user groups. Reliability bias is a particular concern for on-device ML as it counter-acts the promise of technology-enabled service access, an important value proposition of on-device ML. If reliability bias remains unidentified and is not accounted for, it can be a source of unfairness in on-device ML systems. Unfair on-device ML systems that are deployed at scale can lead to a discriminatory service infrastructure that restricts who has access to services, and how these services can be accessed.

Our empirical study shows that design choices made by engineers in each stage of the on-device ML workflow can introduce reliability bias when deploying ML as a system component. While bias in on-device ML can be cast as an AI ethics concern, we consider it important to also approach it as a matter of responsible design. Based on our findings we do not consider reliability bias as an immutable property of a particular model or system. Instead, we position that reliability bias arises from design choices that amplify or reduce disparate predictive performance across groups of users based on their personal attributes. Engineers thus have an active role to play to detect and mitigate reliability bias. While some design choices may lie beyond the immediate control of engineers, they have full control over others. Measuring bias in the on-device ML development workflow is the first step that engineers should take to practice responsible design and make a commitment to building fairer technology systems. 

We have focused our evaluation of reliability bias on performance discrepancies in predictive accuracy. In on-device ML applications, system efficiency is another important performance metric that interacts with accuracy. For example, a KWS system with poor predictive performance can require several user attempts to activate the system. This can increase computations, which leads to increased power consumption and faster drainage of a device's battery. Reliability bias should thus also be considered for system efficiency. The bias measure that we have proposed can be extended easily to characterize reliability bias due to system (in)efficiency. We will investigate this in future work. Additionally, we note that our empirical study is focused on audio-based ML. Although audio is a prominent data modality in on-device ML, we are cognizant that other data types (e.g. images) are also used. Future work can extend our methodology to different modalities and new learning tasks to investigate reliability bias in them.

\subsection{Limitations}
In quantifying an unobservable, abstract construct like fairness, bias measures often make assumptions about what is fair. Yet, fairness is a contested construct~\cite{Jacobs2021Measurement} that is underpinned by the values of those that define it. This makes it important to state assumptions explicitly to avoid mismatches between the construct that is measured, and its quantified operationalization. In this study we have made the assumption that false positive and false negative error rates are equally important across all keyword classes. We have operationalized this assumption by using the Matthews Correlation Coefficient (MCC) to quantify reliability bias. While the MCC is an accepted metric for multiclass classification, it does not capture the disparate impact that false positives and false negatives may carry in particular application scenarios. For example, a KWS system in an emergency care application that has a high false negative rate for the keyword "help" is likely to have a more detrimental impact on affected users than a home entertainment system with a high false positive rate for the keyword "lights on". Characterising harms associated with applications and identifying acceptable error rates is an important area for future research. 

We motivated our use of a parity-based bias measure by claiming that ground truth labels in KWS are exactly known and undisputed. While this avoids bias propagation through labelling choices, constructing groups remains a normative design decision that requires careful consideration. In our audio KWS study we constructed groups based on a speaker's gender. Our approach to labelling voice samples with gender was limited to a binary gender classification system and a crowd-sourced labelling campaign. Even though the MSWC gender labels are self-annotated, binary gender representation removes individuals that do not fit within this classification system from the bias evaluation. Crowd-sourced labelling can introduce further misclassification~\cite{Saito2021vocalturk}. Male and female voices can be higher or lower pitched than what a data worker perceives as normal for that gender, and misclassified accordingly. Gender is also just one of many demographic attributes that influences the human voice~\cite{Singh2019Profiling}. Subgroups established along other speaker attributes can reveal further dimensions of bias and should be investigated in future work.

While dataset representation was not the focus of this study, it oftentimes is an important contributor to bias. We took this into consideration and carefully constructed gender-balanced dataset splits when we designed our experiments. Our gender-balancing protocol resulted in balanced datasets across keyword-speaker pairs (see Table~\ref{tab:mswc_splits}) but unequal utterances/keyword across dataset splits and genders. When the number of unique speakers and utterances/keyword in a dataset are small, it becomes difficult to construct representative datasets, which can affect the reliability of results. In this study, our dataset construction choices may explain some of the performance deviations we observed for the MSWC Kinyarwanda dataset (see Figure~\ref{fig:pruninghps_dataset-lr}), which contained an order of magnitude fewer different speakers than the other datasets. Finding ways for creating balanced datasets and evaluating bias when data availability across groups is variable remains an important open challenge.

We note that our study is limited to investigating bias in CNN architectures. These architectures are very popular for speech and vision related tasks in on-device ML. We chose to focus on one architecture to study how light-weight architectures, a model optimization design choice to reduce the model size, impacts reliability bias. Future work should also examine reliability bias in other architectures. Furthermore, while this study investigates the design choices that we deemed most likely to impact bias, future work should examine the impact of design choices that we did not examine, such as quantization. Despite these limitations, our investigation of performance disparity provides necessary insights that highlight the need of addressing bias in on-device settings. Studying bias in the emerging field of on-device ML is thus an important research direction for the software engineering community. 

%% file: sections/8_conclusion.tex
\section{Conclusion}
\label{s:conclusion}

Billions of devices deploy on-device ML today. Despite bias and fairness being a major area of concern in machine learning (ML), they have not been considered in on-device ML settings. Biased performance impacts device reliability, and can result in systematic device failures due to performance disparities across user groups. This can inconvenience and even harm users. Our study is the first investigation of bias in development workflows in the emerging on-device ML domain, and lays an important foundation for building fairer on-device ML systems. 

In this study we investigate the propagation of bias through design choices in the on-device ML workflow, and identify \emph{reliability bias} as a potential source of unfairness. \emph{Reliability bias} arises from disparate on-device ML performance due to demographic attributes of users, and results in systematic device failure across user groups. Drawing on definitions of group fairness, we quantify \emph{reliability bias} and use the measure in empirical experiments to evaluate the impact of design choices on bias in an audio keyword spotting (KWS) task, a dominant application of on-device ML. Our results validate that seemingly innocuous design choices~--~a light-weight architecture, the data sample rate, pre-processing parameters of input features, and pruning hyper-parameters for model compression~--~can result in disparate predictive performance across male and female groups. 

Given their context dependence and ubiquitous nature, developing inclusive on-device ML systems ought to be an important priority for engineers. Our findings caution that design choices in the development workflow can have major consequences for the propagation of \emph{reliability bias} and consequently fairness of on-device ML. Based on our findings, we suggest strategies for model selection and targeted experimentation to help engineers navigate the gap between technical choices, deployment constraints, accuracy and bias during on-device ML development. Taken together, our work highlights that engineers and the decisions they make have an important role to play to ensure that the social requirement of inclusive on-device ML is realized within the constrained on-device setting.

%% file: sections/appendix.tex
\section{Appendix}
\label{appendix}

\subsection{Experiment Setup: Datasets}

\noindent
\textbf{Google Speech Commands keyword classes:}
{'bed':0, 'bird':1, 'cat':2, 'dog':3, 'down':4, 'eight':5, 'five':6, 'four':7, 'go':8, 'happy':9, 'house':10, 'left':11, 'marvin':12, 'nine':13, 'no':14, 'off':15, 'on':16, 'one':17, 'right':18, 'seven':19, 'sheila':20, 'six':21, 'learn':22, 'stop':23, 'three':24, 'tree':25, 'two':26, 'up':27, 'wow':28, 'yes':29, 'zero':30, 'backward':31, 'follow':32, 'forward':33, 'visual':34}

\parjump{}
\noindent
\textbf{MSWC English keyword classes:}
'about': 0, 'after': 1, 'also': 2, 'been': 3, 'could': 4, 'first': 5, 'from': 6, 'have': 7, 'however': 8, 'just': 9, 'know': 10, 'like': 11, 'many': 12, 'more': 13, 'most': 14, 'only': 15, 'other': 16, 'over': 17, 'people': 18, 'said': 19, 'school': 20, 'some': 21, 'that': 22, 'they': 23, 'this': 24, 'three': 25, 'time': 26, 'used': 27, 'were': 28, 'what': 29, 'when': 30, 'will': 31, 'with': 32, 'would': 33, 'your': 34

\parjump{}
\noindent
\textbf{MSWC German keyword classes:}
'aber': 0, 'alle': 1, 'auch': 2, 'dann': 3, 'dass': 4, 'diese': 5, 'doch': 6, 'durch': 7, 'eine': 8, 'gibt': 9, 'haben': 10, 'hauptstadt': 11, 'heute': 12, 'hier': 13, 'immer': 14, 'jetzt': 15, 'kann': 16, 'können': 17, 'mehr': 18, 'muss': 19, 'nach': 20, 'nicht': 21, 'noch': 22, 'oder': 23, 'schon': 24, 'sein': 25, 'sich': 26, 'sind': 27, 'wenn': 28, 'werden': 29, 'wieder': 30, 'wird': 31, 'wurde': 32, 'zwei': 33, 'über': 34

\parjump{}
\noindent
\textbf{MSWC French keyword classes:}
'alors': 0, 'aussi': 1, 'avec': 2, 'bien': 3, 'cent': 4, 'cette': 5, 'comme': 6, 'c’est': 7, 'dans': 8, 'deux': 9, 'donc': 10, 'elle': 11, 'fait': 12, 'huit': 13, 'mais': 14, 'mille': 15, 'monsieur': 16, 'même': 17, 'nous': 18, 'numéro': 19, 'plus': 20, 'pour': 21, 'quatre': 22, 'saint': 23, 'sept': 24, 'soixante': 25, 'sont': 26, 'tout': 27, 'trois': 28, 'très': 29, 'vingt': 30, 'vous': 31, 'également': 32, 'était': 33, 'être': 34

\parjump{}
\noindent
\textbf{MSWC Kinyarwanda keyword classes:}
'abantu': 0, 'ariko': 1, 'avuga': 2, 'bari': 3, 'benshi': 4, 'buryo': 5, 'cyane': 6, 'gihe': 7, 'gukora': 8, 'gusa': 9, 'hari': 10, 'ibyo': 11, 'icyo': 12, 'igihe': 13, 'imana': 14, 'imbere': 15, 'kandi': 16, 'kuba': 17, 'kugira': 18, 'kuko': 19, 'kuri': 20, 'mbere': 21, 'muri': 22, 'ndetse': 23, 'neza': 24, 'ntabwo': 25, 'nyuma': 26, 'perezida': 27, 'rwanda': 28, 'ubwo': 29, 'umuntu': 30, 'umwe': 31, 'yagize': 32, 'yari': 33, 'yavuze': 34

\newpage
\subsection{Statistical Analysis: Design Choices Arising During Model Training}

\lstset{basicstyle=\footnotesize\ttfamily,breaklines=true,showstringspaces=false}
\renewcommand{\lstlistingname}{Model}
\begin{lstlisting}[language=python,caption=First factorial ANOVA interaction model for model training design choices,label=mod:anova1]
model_inital = 'metric ~ C(dataset_name, Sum)+C(model_arch, Sum)C(resample_rate, Sum)+C(mfccs, Sum)+C(mel_bins, Sum)+C(frame_length, Sum)+C(frame_step, Sum)+C(window_fn, Sum)+C(dataset_name, Sum)*C(model_arch, Sum)*C(resample_rate, Sum)*C(mfccs, Sum)*C(mel_bins, Sum)+C(dataset_name, Sum)*C(model_arch, Sum)*C(resample_rate, Sum)*C(frame_length, Sum)*C(frame_step, Sum)*C(window_fn, Sum)'
\end{lstlisting}

\lstset{basicstyle=\footnotesize\ttfamily,breaklines=true,showstringspaces=false}
\renewcommand{\lstlistingname}{Model}
\begin{lstlisting}[language=python,caption=Final factorial ANOVA interaction model for effect of model training design choices on MCC,label=mod:anova2]
model_final_mcc = 'mcc ~ C(dataset_name, Sum)+C(model_arch, Sum)+C(resample_rate, Sum)+C(mfccs, Sum)+C(mel_bins, Sum)+C(dataset_name, Sum)*C(resample_rate, Sum)*C(mfccs, Sum)+C(model_arch, Sum)*C(mfccs, Sum)*C(mel_bins, Sum)+C(dataset_name, Sum)*C(model_arch, Sum)*C(mfccs, Sum)+C(frame_length, Sum)+C(frame_step, Sum)+C(model_arch, Sum)*C(frame_length, Sum)*C(frame_step, Sum)'
\end{lstlisting}

\lstset{basicstyle=\footnotesize\ttfamily,breaklines=true,showstringspaces=false}
\renewcommand{\lstlistingname}{Model}
\begin{lstlisting}[language=python,caption=Final factorial ANOVA interaction model for effect of model training design choices on reliability bias,label=mod:anova3]
model_final_bias = 'reliability_bias ~ C(dataset_name, Sum)+C(model_arch, Sum)+C(resample_rate, Sum)+C(dataset_name, Sum)*C(resample_rate, Sum)+C(mfccs, Sum)+C(mel_bins, Sum)+C(dataset_name, Sum)*C(model_arch, Sum)*C(mfccs, Sum)+C(dataset_name, Sum)*C(mel_bins, Sum)+C(frame_length, Sum)'
\end{lstlisting}

\newpage
\subsection{Impact of Pre-processing parameters}

\begin{figure}[!hbt]
        \centering
        \includegraphics[width=\textwidth]{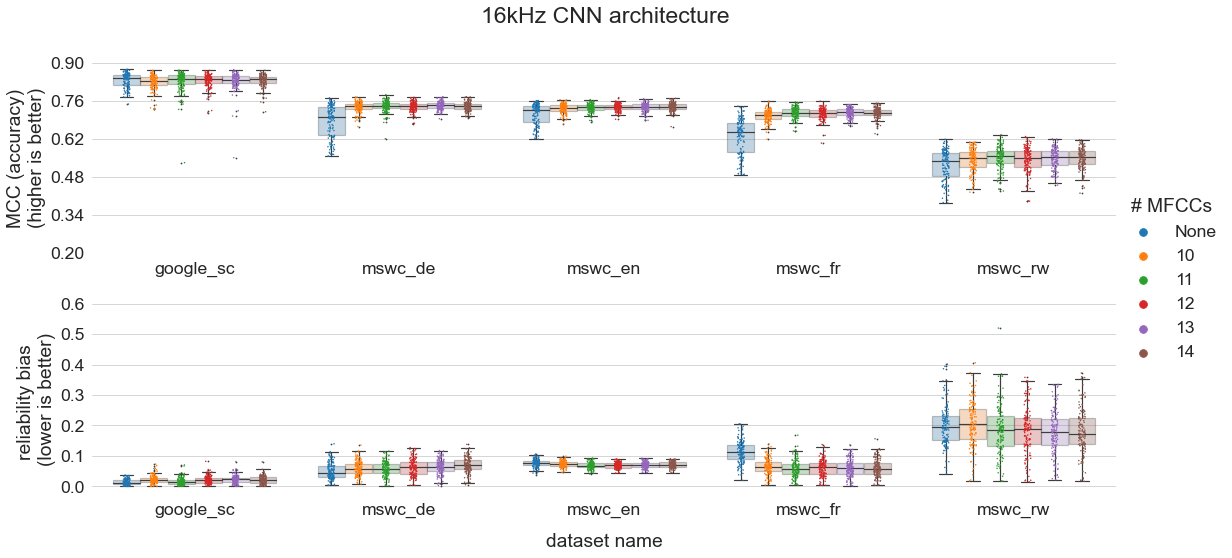}
    \caption{Effect of MFCC dimensions on accuracy and reliability bias for 16kHz CNN models. Models without MFCC features, i.e. models that directly use log Mel spectrograms as input features, tend to perform worse than those that use MFCC features.}
\label{fig:dataset_arch_mfccs_16}
\end{figure}

\begin{figure}[hbt]
        \centering
        \includegraphics[width=\textwidth]{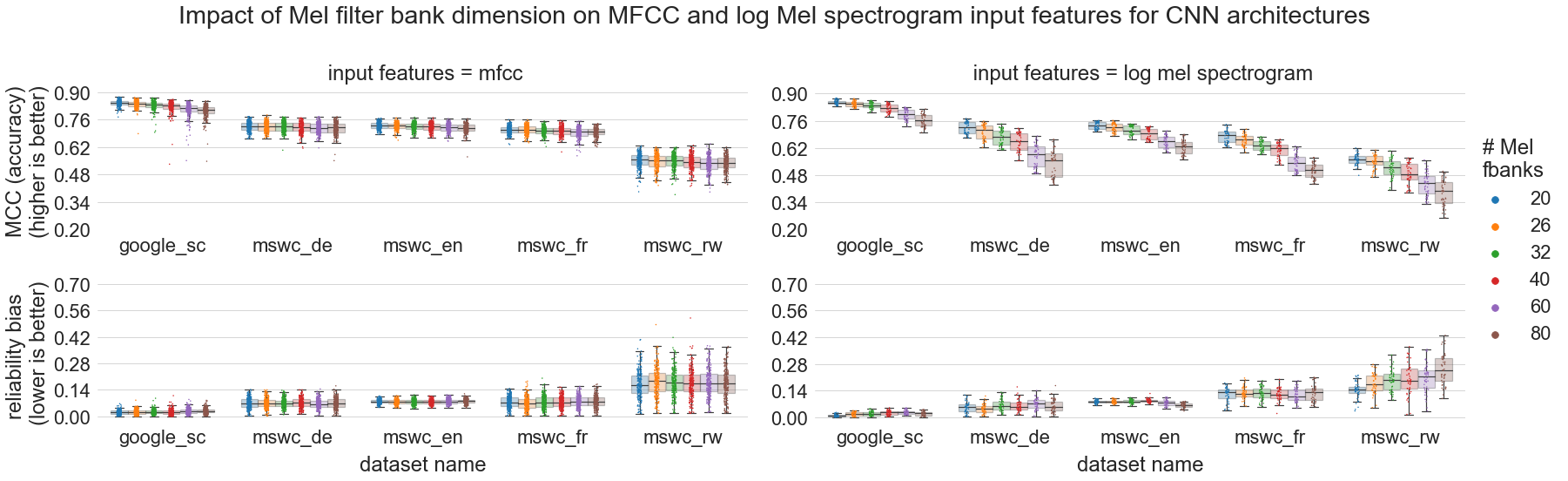}
    \caption{Effect of log Mel spectrogram dimensions (\# Mel fbanks) on accuracy and reliability bias, disaggregated by input feature type for CNN architectures. The number of Mel filter banks clearly impacts models that directly use log Mel spectrograms as input features.}
\label{fig:dataset_arch_melbins_feature_type_cnn}
\end{figure}

\newpage
\subsection{Statistical Analysis: Design Choices Arising During Model Optimization}

\lstset{basicstyle=\footnotesize\ttfamily,breaklines=true,showstringspaces=false}
\renewcommand{\lstlistingname}{Model}
\begin{lstlisting}[language=python,caption=First factorial ANOVA interaction model for pruning hyper-parameters,label=mod:anova4]
model_inital = 'delta_metric ~ mcc_baseline + reliability_bias_baseline + C(dataset_name, Sum)+C(model_arch, Sum)+C(resample_rate, Sum)+C(pruning_learning_rate, Sum)+C(pruning_schedule, Sum)+C(pruning_frequency, Sum)+C(pruning_final_sparsity, Sum)+C(dataset_name, Sum)*C(model_arch, Sum)*C(resample_rate, Sum)*C(pruning_learning_rate, Sum)*C(pruning_schedule, Sum)*C(pruning_frequency, Sum)*C(pruning_final_sparsity, Sum)'
\end{lstlisting}

\lstset{basicstyle=\footnotesize\ttfamily,breaklines=true,showstringspaces=false}
\renewcommand{\lstlistingname}{Model}
\begin{lstlisting}[language=python,caption=Final factorial ANOVA interaction model for effect of pruning design choices on change in MCC,label=mod:anova5]
model_final_delta_mcc = 'delta_mcc ~ mcc_baseline + reliability_bias_baseline + C(dataset_name, Sum)+C(model_arch, Sum)+C(resample_rate, Sum)+C(pruning_learning_rate, Sum)+C(pruning_schedule, Sum)+C(pruning_frequency, Sum)+C(pruning_final_sparsity, Sum)+C(model_arch, Sum)*C(pruning_learning_rate, Sum)*C(pruning_schedule, Sum)*C(pruning_final_sparsity, Sum)+C(dataset_name, Sum)*C(model_arch, Sum)*C(resample_rate, Sum)*C(pruning_final_sparsity, Sum)+C(dataset_name, Sum)*C(pruning_learning_rate, Sum)*C(pruning_schedule, Sum)+C(dataset_name, Sum)*C(pruning_schedule, Sum)*C(pruning_final_sparsity, Sum)+C(dataset_name, Sum)*C(pruning_learning_rate, Sum)*C(pruning_final_sparsity, Sum)+C(resample_rate, Sum)*C(pruning_learning_rate, Sum)*C(pruning_final_sparsity, Sum)'
\end{lstlisting}

\lstset{basicstyle=\footnotesize\ttfamily,breaklines=true,showstringspaces=false}
\renewcommand{\lstlistingname}{Model}
\begin{lstlisting}[language=python,caption=Final factorial ANOVA interaction model for effect of pruning design choices on change in reliability bias,label=mod:anova6]
model_final_delta_bias = 'delta_reliability_bias ~ mcc_baseline+reliability_bias_baseline+C(dataset_name, Sum)+C(model_arch, Sum)+C(resample_rate, Sum)+C(pruning_learning_rate, Sum)+C(pruning_schedule, Sum)+C(pruning_frequency, Sum)+C(pruning_final_sparsity, Sum)+C(dataset_name, Sum)*C(model_arch, Sum)*C(resample_rate, Sum)*C(pruning_learning_rate, Sum)+C(dataset_name, Sum)*C(pruning_learning_rate, Sum)*C(pruning_final_sparsity, Sum)+C(pruning_schedule, Sum)*C(pruning_final_sparsity, Sum)+C(model_arch, Sum)*C(pruning_final_sparsity, Sum)'
\end{lstlisting}

\newpage
\subsection{Model Selection After Pruning}

\begin{figure}[hbt]
        \centering
        \includegraphics[width=\textwidth]{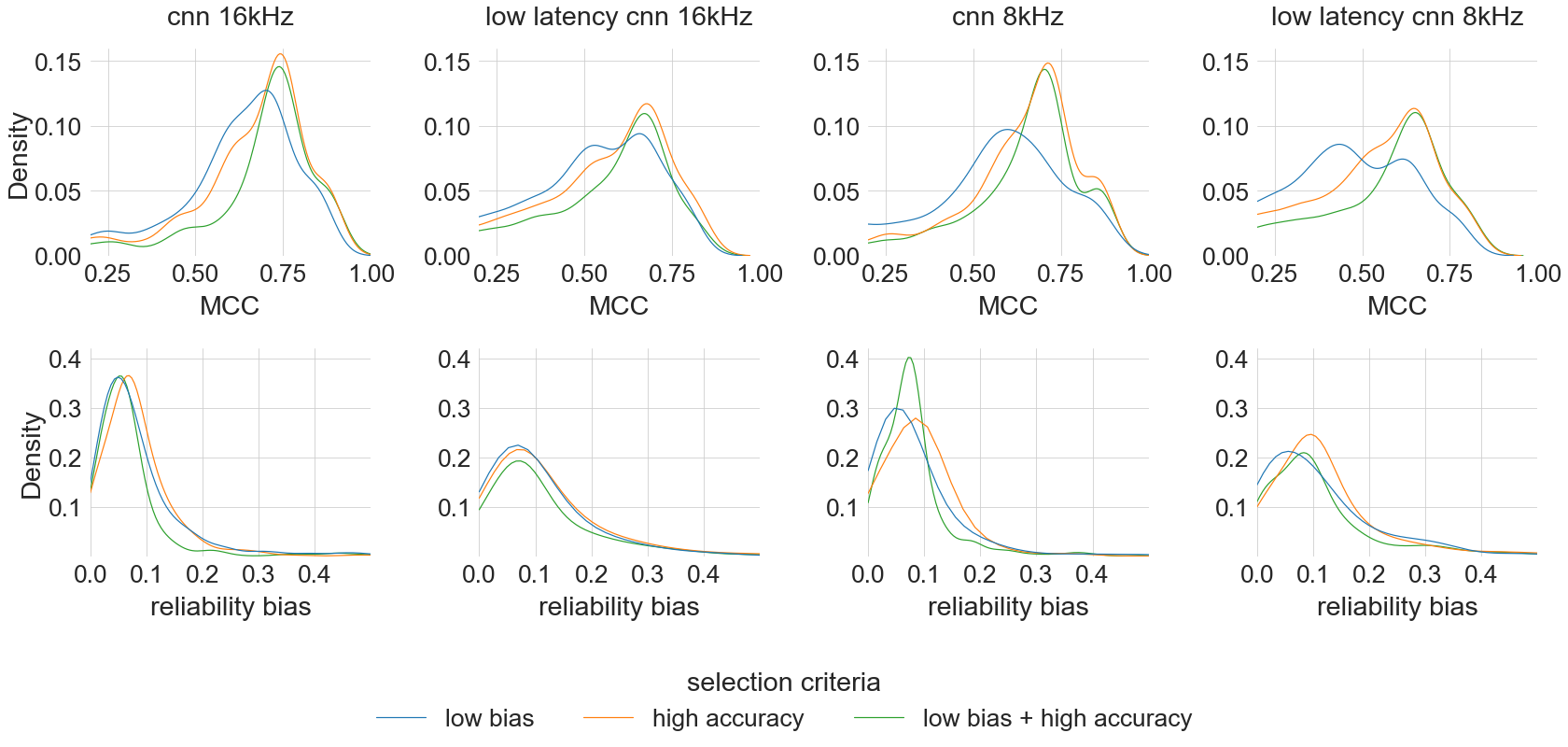}
        \caption{MCC (accuracy) score and reliability bias of models after pruning. Models were selected based on three selection criteria: high accuracy, low bias, and high accuracy + low bias. After pruning MCC is greatest for models selected with a criteria that considers high accuracy. Similarly, reliability bias is lower for criteria that consider low bias.}
        \label{fig:pruninghps_selection_criteria_metric}
\end{figure}